%% file: ms.tex
\documentclass[10pt,twocolumn,letterpaper]{article}

\usepackage{tikz}
\usepackage{iccv_final}
\usepackage{times}
\usepackage{epsfig}
\usepackage{graphicx}
\usepackage{amsmath}
\usepackage{amssymb}
\usepackage{multirow}
\usepackage{tabularx}
\usepackage{array}
\usepackage{color}
\usepackage{colortbl}
\usepackage[nocompress]{cite}

\input{common/Commands}

\usepackage[pagebackref=true,breaklinks=true,letterpaper=true,colorlinks,bookmarks=false]{hyperref}

\iccvfinalcopy 

\def\iccvPaperID{5103} 

\makeatletter
\newcommand\copyrighttext{%
  \footnotesize \copyright 2019 IEEE. Personal use of this material is permitted.
  Permission from IEEE must be obtained for all other uses, in any current or future
  media, including reprinting/republishing this material for advertising or promotional
  purposes, creating new collective works, for resale or redistribution to servers or
  lists, or reuse of any copyrighted component of this work in other works.
  DOI: Pending}

\newcommand\copyrightnotice{%
\begin{tikzpicture}[remember picture,overlay]
\node[anchor=south,yshift=10pt] at (current page.south) {\fbox{\parbox{\dimexpr\textwidth-\fboxsep-\fboxrule\relax}{\copyrighttext}}};
\end{tikzpicture}%
}

\makeatother

\ificcvfinal\fi
\begin{document}

\title{Mono-SF: Multi-View Geometry Meets Single-View Depth \\ for Monocular Scene Flow Estimation of Dynamic Traffic Scenes}

\author{Fabian Brickwedde$^{1,2}$\\
$^1$ Robert Bosch GmbH, Germany\\
{\tt\small Fabian.Brickwedde@de.bosch.com}
\and
Steffen Abraham$^1$\\
$^2$ VSI Lab, Goethe University Frankfurt, Germany \\
First line of institution2 address\\
{\tt\small Steffen.Abraham@de.bosch.com}
\and
Rudolf Mester$^{2,3}$\\
$^3$ NTNU Trondheim, Norway \\
{\tt\small rudolf.mester@ntnu.no}
}

\maketitle
\ificcvfinal\thispagestyle{empty}\fi
\copyrightnotice

\begin{abstract}
\input{text/Abstract/Abstract.tex}
\end{abstract}

\section{Introduction}
\input{text/Introduction/Introduction}
\section{Related work}
\input{text/Introduction/RelatedWork}
\section{Method}
\input{text/Method/SingleImageDepth}
\input{text/Method/MonoSceneFlow}
\section{Experiments}
\input{text/Experiments/Experiments}
\section{Conclusion}
\input{text/Conclusion/Conclusion}

\clearpage
{\small
\bibliographystyle{ieee_fullname}
\bibliography{Literatur/Literatur}
}

\end{document}

%% file: common/Commands.tex
\newcommand\RotText[1]{\rotatebox{90}{\parbox{1.25cm}{\centering#1}}}

\newcommand{\vectorsymbol}[1]{\mathbf{#1}}

\newcommand{\vt}{\vectorsymbol{t}}

\newcommand{\vn}{\vectorsymbol{n}}

\newcommand{\vp}{\vectorsymbol{p}}

\newcommand{\matrixsymbol}[1]{\mathbf{#1}}

\newcommand{\mR}{\matrixsymbol{R}}
\newcommand{\mK}{\matrixsymbol{K}}

\newcommand{\mX}{\matrixsymbol{X}}
\newcommand{\mT}{\matrixsymbol{T}}


%% file: text/Abstract/Abstract.tex
Existing 3D scene flow estimation methods provide the 3D geometry and 3D motion of a scene 
 and gain a lot of interest, for example in the context of autonomous driving.
These methods are traditionally based on a temporal series of stereo images.
In this paper, we propose a novel monocular 3D scene flow estimation method, called Mono-SF.
Mono-SF jointly estimates the 3D structure and motion of the scene 
by combining multi-view geometry and single-view depth information.
Mono-SF considers that the scene flow should be consistent in terms of warping the reference image
in the consecutive image based on the principles of multi-view geometry.
For integrating single-view depth in a statistical manner, a convolutional neural network, called ProbDepthNet, is proposed. 
ProbDepthNet estimates pixel-wise depth distributions from a single image rather than single depth values.
Additionally, as part of ProbDepthNet, a novel recalibration technique for regression problems is proposed to ensure well-calibrated distributions.
Our experiments show that Mono-SF outperforms state-of-the-art monocular baselines
 and ablation studies support the Mono-SF approach and ProbDepthNet design.

%% file: text/Introduction/Introduction.tex
\input{images/Intro/Figure.tex}

In applications such as mobile robots or autonomous vehicles
 a representation of the surrounding environment is utilized,
e.g. to fulfill a navigation task.
From a computer vision point of view, the 3D position and motion
 of a pixel in the image is denoted as \emph{3D scene flow} \cite{vedula1999three,vedula2005three},
 which is traditionally estimated based on a temporal series of stereo images \cite{vogel2013piecewise,Menze2015CVPR,behl2017bounding}.
In this work, we propose a novel scene flow estimation method, \emph{Mono-SF}, for a monocular camera setup focusing on dynamic traffic scenes.
Monocular camera systems are often preferred over stereo cameras due to being more cost efficient and to avoid the effort of calibrating the stereo rig.
However, 3D scene flow estimation is an ill-posed problem in a monocular camera setup.
To solve the ambiguity, previous monocular methods assumed that the moving objects are in contact with the surrounding environment \cite{monostixel,ranftl2016dense, bullinger20183d} 
or that the scene follows a smoothness prior regarding surface and motion \cite{mitiche15mono, Xiao2017, kumar2017monocular}.
These assumptions might be violated and the methods still require a relative translational motion of the camera to the scene. 
In contrast to the multi-view geometry-based approaches, methods were proposed (e.g. \cite{eigen2014depth,godard2016deepdepth,fu2018deep})
that provide depth estimates from a single image at a reasonable level of quality. 
However, single-view depth estimation and multi-view geometry are mostly tackled as two individual tasks 
or fused in a way that is only applicable for static scenes \cite{tateno2017cnn,facil2017fusion,yin2017scale}.
Our proposed Mono-SF method combines \emph{multi-view geometry} with \emph{single-view depth information}
in a probabilistic optimization framework to provide consistent 3D scene flow estimates.
Thereby, both kinds of information are exploited and the single-view depth serves to solve the multi-view geometry-based ambiguity.
 
Previous methods
 \cite{Menze2015CVPR,MENZE201860,behl2017bounding} 
 showed that a suitable representation of particularly traffic scenes
is the decomposition into 3D planar surface elements, each one assigned to a rigid body.
A rigid body is either the background or a potentially moving object. 
Following this model,
 Mono-SF jointly estimates the 3D geometry of each plane and 6D motion of each rigid body
considering 
 a) the multi-view geometry by warping the reference image into the consecutive image, 
 b) probabilistic single-view depth estimates, and
 c) scene model smoothness priors (see Fig. \ref{fig_introduction}). 
Additionally, an instance segmentation is exploited to detect the set of potentially moving objects.

As an additional contribution, we propose \emph{ProbDepthNet}, 
a convolutional neural network (CNN) that estimates pixel-wise probability depth distributions from a single image
 rather than just single depth values such as \cite{eigen2014depth,godard2016deepdepth,fu2018deep}.
Whereas the problem of overconfident estimates is a well-known problem in classification \cite{guo2017calibration},
it is typically ignored in probabilistic approaches for regression \cite{ICKMB18,gast2018lightweight,kendall2017uncertainties,klodt2018supervising}.
Therefore, we propose a novel \emph{recalibration technique}:
CalibNet, a small subsequent part of ProbDepthNet, is trained on a hold-out split of the training data
to compensate for overfitting effects and to provide well-calibrated distributions.

Our Mono-SF approach is evaluated with respect to several state-of-the-art monocular baselines and
an ablation study confirms the importance of the individual components of the proposed optimization framework.
Furthermore, ProbDepthNet is validated to provide well-calibrated depth distributions.
Our experiments show that several previous probabilistic approaches suffer from overconfident estimates --
 an effect that could be compensated by adding our proposed CalibNet for recalibration.
The suitability of ProbDepthNet for integrating single-view depth information in Mono-SF is confirmed,
 especially due to the importance of providing single-view depth information in a probabilistic and well-calibrated form.

%% file: images/Intro/Figure.tex
\begin{figure}[htbp]
\centering
	\centering
	\def\svgwidth{0.48\textwidth}
	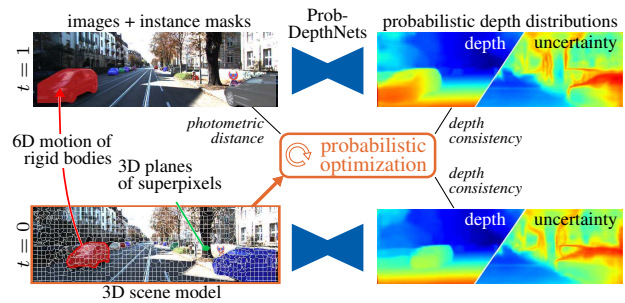
  \caption{Overview of Mono-SF for monocular scene flow estimation. 
    Mono-SF jointly optimizes the 3D geometry of a set of planes with the 6D motion of rigid bodies
	considering a) a photometric distance by warping the reference image into the consecutive image, b)
	probabilistic depth distributions provided by ProbDepthNet and c) scene model smoothness priors.}
	\label{fig_introduction}
\end{figure}

%% file: images/Intro/IntroVersion7.eps_tex
\begingroup%
  \makeatletter%
  \providecommand\color[2][]{%
    \errmessage{(Inkscape) Color is used for the text in Inkscape, but the package 'color.sty' is not loaded}%
    \renewcommand\color[2][]{}%
  }%
  \providecommand\transparent[1]{%
    \errmessage{(Inkscape) Transparency is used (non-zero) for the text in Inkscape, but the package 'transparent.sty' is not loaded}%
    \renewcommand\transparent[1]{}%
  }%
  \providecommand\rotatebox[2]{#2}%
  \newcommand*\fsize{\dimexpr\f@size pt\relax}%
  \newcommand*\lineheight[1]{\fontsize{\fsize}{#1\fsize}\selectfont}%
  \ifx\svgwidth\undefined%
    \setlength{\unitlength}{257.55000655bp}%
    \ifx\svgscale\undefined%
      \relax%
    \else%
      \setlength{\unitlength}{\unitlength * \real{\svgscale}}%
    \fi%
  \else%
    \setlength{\unitlength}{\svgwidth}%
  \fi%
  \global\let\svgwidth\undefined%
  \global\let\svgscale\undefined%
  \makeatother%
  \begin{picture}(1,0.50496605)%
    \scriptsize
    \lineheight{1}%
    \setlength\tabcolsep{0pt}%
    \put(0.04,0.052){\includegraphics[width=0.94\unitlength]{images/Intro/IntroVersion7.eps}}%
    \put(0.005,0.27){\color[rgb]{0,0,0}\makebox(0,0)[lt]{\lineheight{0.78125}\smash{\begin{tabular}[t]{r}\cellcolor{white} 6D motion of\\\cellcolor{white} rigid bodies \end{tabular}}}}%
    \put(0.18,0.23){\color[rgb]{0,0,0}\makebox(0,0)[lt]{\lineheight{0.78125006}\smash{\begin{tabular}[t]{l}3D planes \\ of superpixels\end{tabular}}}}%
    \put(0.45,0.47){\color[rgb]{0,0,0}\makebox(0,0)[lt]{\lineheight{0.78125006}\smash{\begin{tabular}[t]{c}Prob-\\DepthNets\end{tabular}}}}%
    \put(0.50052148,0.26249254){\color[rgb]{0.89019608,0.41960784,0.19215686}\makebox(0,0)[lt]{\lineheight{0.78125006}\smash{\begin{tabular}[t]{c}\footnotesize probabilistic\\ \footnotesize optimization\end{tabular}}}}%
    \put(0.6,0.46){\color[rgb]{0,0,0}\makebox(0,0)[lt]{\lineheight{0.78125006}\smash{\begin{tabular}[t]{l}probabilistic depth distributions\end{tabular}}}}%
    \put(0.705,0.30){\color[rgb]{0,0,0}\makebox(0,0)[lt]{\lineheight{0.78125006}\smash{\begin{tabular}[t]{l}\fontsize{6}{13}\selectfont\textit{depth}\\\fontsize{6}{13}\selectfont \textit{consistency}\end{tabular}}}}%
    \put(0.705,0.22){\color[rgb]{0,0,0}\makebox(0,0)[lt]{\lineheight{0.78125006}\smash{\begin{tabular}[t]{l}\fontsize{6}{13}\selectfont \textit{depth}\\ \fontsize{6}{13}\selectfont\textit{consistency}\end{tabular}}}}%
    \put(0.29,0.3){\color[rgb]{0,0,0}\makebox(0,0)[lt]{\lineheight{0.78125006}\smash{\begin{tabular}[t]{r}\fontsize{6}{13}\selectfont \textit{photometric} \\ \fontsize{6}{13}\selectfont \textit{distance}\end{tabular}}}}%
    \put(0.1,0.46){\color[rgb]{0,0,0}\makebox(0,0)[lt]{\lineheight{0.78125006}\smash{\begin{tabular}[t]{l}images + instance masks\end{tabular}}}}%
    \put(0.155,0.03){\color[rgb]{0,0,0}\makebox(0,0)[lt]{\lineheight{0.78125006}\smash{\begin{tabular}[t]{l}3D scene model\end{tabular}}}}%
    \put(0.73,0.42490028){\color[rgb]{1,1,1}\makebox(0,0)[lt]{\lineheight{0.78125006}\smash{\begin{tabular}[t]{l}depth\end{tabular}}}}%
    \put(0.84,0.4264446){\color[rgb]{0,0,0}\makebox(0,0)[lt]{\lineheight{0.78125006}\smash{\begin{tabular}[t]{l}uncertainty\end{tabular}}}}%
    \put(0.84,0.14638363){\color[rgb]{0,0,0}\makebox(0,0)[lt]{\lineheight{0.78125006}\smash{\begin{tabular}[t]{l}uncertainty\end{tabular}}}}%
    \put(0.73,0.14611345){\color[rgb]{1,1,1}\makebox(0,0)[lt]{\lineheight{0.78125006}\smash{\begin{tabular}[t]{l}depth\end{tabular}}}}%
	\put(0.02,0.04){\color[rgb]{0,0,0}\makebox(0,0)[lt]{\lineheight{0.78125006}\smash{\begin{tabular}[t]{l}\RotText{$t=0$}\end{tabular}}}}%
	\put(0.02,0.32){\color[rgb]{0,0,0}\makebox(0,0)[lt]{\lineheight{0.78125006}\smash{\begin{tabular}[t]{l}\RotText{$t=1$}\end{tabular}}}}%
  \end{picture}%
\endgroup%

%% file: text/Introduction/RelatedWork.tex
The works related to the approach presented here are divided into three categories:
In the first category are the stereo-based scene flow methods which inspired
 our Mono-SF scene model and optimization framework.
The second category provides an overview of methods for monocular scene reconstruction
comprising the baseline methods.
Finally, the category of probabilistic deep learning represents works 
 related to the probabilistic design of ProbDepthNet.

\textbf{Stereo Scene Flow:}
Scene flow estimation
 was introduced by Vedula et al. 
    \cite{vedula1999three, vedula2005three}
as a joint optimization of 3D geometry and motion of the scene
based on a sequence of stereo images. 
Mostly variational approaches were used subsequently to extend the scene flow concept
    \cite{huguet2007variational, pons2007multi, wedel2008efficient, wedel2011stereoscopic, valgaerts2010joint, basha2013multi, herbst2013rgbdflow}.
However,  Vogel et al. 
    \cite{vogel2013piecewise}
were the first that significantly outperformed individual stereo 
and optical flow methods on their respective tasks for dynamic traffic scenes. 
They represented the dynamic scene as a collection of rigid moving planar surface elements
 and jointly optimized the geometry and the motion of each plane considering scene model priors. 
Menze et al. 
    \cite{Menze2015CVPR}
formulated the problem by a set of rigid moving objects
and jointly optimized their motion with the geometry of each plane. 
This representation is particularly beneficial if the association of planes to objects is supported by an instance segmentation 
    as proposed in \cite{behl2017bounding}.
Our Mono-SF model corresponds to these approaches, 
called \emph{object} \cite{Menze2015CVPR} or \emph{instance scene flow} \cite{behl2017bounding},
 but Mono-SF uses only monocular images. 

\textbf{Monocular Scene Reconstruction:} 
Traditionally, monocular scene reconstruction is based on the \emph{structure from motion} (SfM) principle.  
The SfM-based approaches can be divided into several categories:
First, rigid SfM-based methods 
estimate the 3D geometry of a rigid scene based on its relative motion to the camera, 
e.g. a static scene and a moving camera
    \cite{mur2015orb, engel2014lsd, fananipmo, pereira2017monocular, ummenhofer2017demon}. 
Second, the non-rigid SfM principle is typically used 
to derive the deformation of a single object 
    \cite{brikbeck2010depth, garg2013dense, golyanik2016nrsfm}. 
Third, multi-body SfM is the concept of 
reconstructing individual moving parts of the scene separately 
    \cite{ranftl2016dense, kumar2017monocular}. 
However, the absolute and relative scales of the reconstructions are unknown in general. 
Scene model assumptions are needed to solve this scale ambiguity, e.g.\ 
that moving objects are in contact with the surrounding environment
    \cite{monostixel,ranftl2016dense, bullinger20183d} 
 or that the scene follows a smoothness prior regarding surface and motion
    \cite{mitiche15mono, Xiao2017, kumar2017monocular}. 

Even though the idea of \emph{single-view depth estimation} is by far not new
    \cite{saxena2005learning,ladicky2014pulling,hoiem2005automatic},
the real breakthrough was achieved by usage of deep learning methods. 
Pioneering, Eigen et al. 
    \cite{eigen2014depth}
proposed a CNN that is trained in a supervised manner and estimates the depth
 in a coarse to fine scheme. 
Afterward, various self-supervised and unsupervised approaches were proposed
using either an image reconstruction loss in a stereo setup
    \cite{garg2016singleviewdepth,godard2016deepdepth}
or in a monocular image sequence
    \cite{zhan2018unsupervised,wang2018learning,mahjourian2018unsupervised,zhou2017unsupervised}. 
Fu et al. 
    \cite{fu2018deep}
formulated the depth estimation as an ordinal regression problem,
 which led to the currently leading approach in the KITTI depth prediction benchmark
    as reported by \cite{uhrig2017sparsity}.
Multi-task CNNs that estimate optical flow alongside the depth were proposed
    \cite{yin2018geonet,zou2018df,yang2018every,teng2018occlusion}.
Thereby, both tasks benefit from each other by a combined training loss.
DeMoN \cite{ummenhofer2017demon} could also exploit multi-view information for depth estimation during inference.
However, it is focused and applied only to static scenes as it just estimates a single camera motion for the whole scene.

Whereas single-view depth estimation and multi-view geometry are mostly taken as individual tasks,
a few works combine both. The single-view depth estimation can be useful for scale estimation in monocular visual odometry
 \cite{barnes2018driven,yin2017scale,yang2018deep}
 or fused with SfM-based depth estimates in static environments
 \cite{tateno2017cnn,facil2017fusion,yin2017scale}. 
Kumar et al. \cite{kumar2019motion} used single-view depth estimation for depth initialization 
 in a multi-body or non-rigid SfM-based approach similar to \cite{kumar2017monocular}.
Brickwedde et al. \cite{monostixel2} proposed a fusion of single-view depth estimates
and optical flow to provide a column-wise segmentation in stick-like rigid elements of particularly traffic scenes.
In contrast to these methods, Mono-SF is formulated as a scene flow estimation problem
and integrates probabilistic single-view depth distributions instead of single depth values.

\textbf{Probabilistic Deep Learning:}
The methods of single-view depth estimation mentioned in the previous section
 do not provide an uncertainty measure or probabilistic distribution of the depth estimates. 
Kendall and Gal
    \cite{kendall2017uncertainties}
distinguished two kind of uncertainties, \emph{epistemic} and \emph{aleatoric uncertainty}. 
Epistemic uncertainty corresponds to the uncertainty of the model parameters
 or the ignorance which model generates the training data, 
 whereas aleatoric uncertainty refers to noise in the input data \cite{kendall2017uncertainties}. 
Malinin et al. \cite{malinin2018predictive} extended this definition by introducing
the \emph{distributional} uncertainty to represent out-of-distribution data.
To estimate the extent of aleatoric uncertainty in a regression problem,
 different strategies have been proposed. 
First, a probability distribution can be learned by minimizing 
 the negative log-likelihood on the training data
    \cite{kendall2017uncertainties,klodt2018supervising}. 
Second, Ilg et al. 
    \cite{ICKMB18}
proposed a single network that is pushed to estimate a complementary set of hypotheses.
Thereby, the aleatoric uncertainty is encoded by the empirical distribution of these hypotheses. 
Third, Gast and Roth
    \cite{gast2018lightweight}
replaced each layer with a probabilistic layer to propagate an input uncertainty through the network. 
The ProbDepthNet method presented here falls under the category of 
estimating the aleatoric uncertainty with a single network and single inference such as
    \cite{ICKMB18,gast2018lightweight,kendall2017uncertainties,klodt2018supervising}. 
For classification problems, Guo et al.
    \cite{guo2017calibration}
 showed that modern neural networks tend to overfit on the training data,
which results in highly overconfident estimates. 
Recalibration techniques were proposed to compensate for this effect \cite{platt1999probabilistic,guo2017calibration,kuleshov2018accurate}.
%

%% file: text/Method/SingleImageDepth.tex
The monocular scene flow estimation method, \emph{Mono-SF}, is designed to
 combine multi-view geometry with probabilistic single-view depth information in a probabilistic optimization framework.
First, a CNN, called \emph{ProbDepthNet}, providing single-view depth information in a probabilistic
 and well-calibrated form is described.
Second, the Mono-SF model and optimization framework are presented.
\subsection{Probabilistic Single-View Depth Estimation}
\label{sec_psid}
To integrate the single-view depth estimates in Mono-SF in a statistical manner,
 ProbDepthNet is designed to represent the uncertainty of each estimate.
Thus, the main objective of ProbDepthNet is not to provide a single depth estimate,
 but to provide a probability density function of the depth for each pixel $\vp$ given an input image $I$. 
The depth is encoded by its inverse form $d = Z^{-1}$,  where $Z$ is the z-coordinate of the 3D-position in camera coordinates. 
ProbDepthNet estimates a pixel-wise probability density function 
 $p_{\vp}(d \mid I)$
 parameterized as a \emph{mixture of Gaussians}:  
\begin{equation}
    p_{\vp}(d \mid I) = \sum_{i=1}^{K} \lambda_i \cdot \mathcal{N} \left(d - \mu_i, \sigma_i \right )
\end{equation}
$K$ represents the number of components, $\lambda_i$ are the weights, 
$\mu_i$ are the mean values, and $\sigma_i$ are the variances of the $i$-th component. 
Compared to a single Gaussian distribution, a mixture model is able to capture more general distributions, e.g. a multimodal distribution.
But, the mixture of Gaussians is more an exemplary choice and other parameterizations of a probability
 distribution can be used as well.  

 \input{images/Architecture/Figure.tex}
Fig. \ref{fig_architecture} gives an overview of the architecture, training process and ground truth generation. 
ProbDepthNet consists of two parts: 
DepthNet and CalibNet. 
DepthNet is a fully convolutional ResNet-50
	\cite{he2016deep}
with skip connections between corresponding encoder and decoder layers. 
The outputs of DepthNet are the parameters of the mixture of Gaussians,
whereby the variance is provided in the log-space $s_i = \log \sigma_i$. 
Additionally, the variances $s_i$ and weights $\lambda_i$ of DepthNet
 are recalibrated by CalibNet, which outputs the corresponding recalibrated values $\tilde{s}_i$ and $\tilde{\lambda}_i$. 
CalibNet just consists of five $1 \times 1$ convolutional layers:
One layer without non-linear activation function to provide a scaled version of the inputs
 and a residual path with four layers including exponential linear units as activation functions. 
The number of features of all layers is equal to the number of inputs $2K$. 
Both networks are trained on different splits of the training data to avoid overfitting of DepthNet on the calibration split.
The negative log-likelihood loss $\mathcal{L}$ is minimized during training similar to \cite{kendall2017uncertainties,klodt2018supervising}:
\begin{equation}
    \mathcal{L} = \sum_{u,v \in \Omega_{GT}} \left[ 
    	- \log \left( \sum_{i=1}^{K} \lambda_i \mathcal{N} \left(d_{GT} - \mu_i, \sigma_i \right ) \right ) \right]
\end{equation}
$u,v \in \Omega_{GT}$ are all pixels in the image with valid ground truth depth values $d_{GT}$ and
$\mu_i,\lambda_i,\sigma_i$ are the outputs of the trained network.

To overcome the limitations of lidar data in terms of density, range, and field of view,
an intermediate fusion based on stereo images is used for ground truth depth generation.
First, the lidar point cloud is projected to the image
and inconsistent measurements are removed to handle occlusion problems. 
Second, these sparse depth maps are completed considering a photometric distance 
 between the two stereo images by using an SGM-based approach
    \cite{hirschmuller2005accurate}. 

ProbDepthNet learns to estimate a pixel-wise depth distribution by observing the depth distribution during the training process.
Thereby, the depth distribution captures the aleatoric uncertainty regarding the theory of Kendall and Gal \cite{kendall2017uncertainties}.
The aleatoric uncertainty is considered to be the most dominant uncertainty in many vision applications \cite{kendall2017uncertainties}.
Our experiments show that CalibNet for recalibration is also applicable to different probabilistic approaches similar to \cite{ICKMB18,gast2018lightweight}.

%% file: images/Architecture/Figure.tex
\begin{figure}[t]
\centering
  \centering
	\def\svgwidth{0.48\textwidth}
	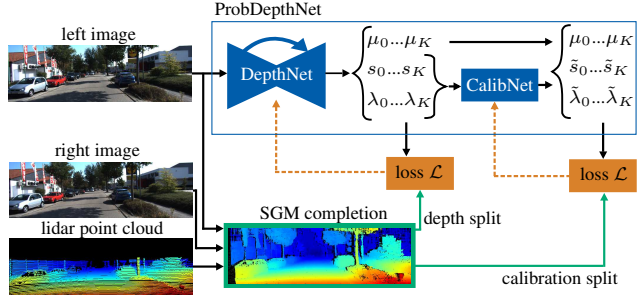
	\caption{Overview of ProbDepthNet for probabilistic single-view depth estimation. 
	  The architecture consists of two parts: DepthNet and CalibNet for recalibration (blue). 
	   Both parts provide a parametrized form ($\mu_i$, $s_i$ / $\tilde{s}_i$ and $\lambda_i$ / $\tilde{\lambda}_i$) of a mixture of Gaussians.
	   Each part is trained on a different split of the training data using a negative log-likelihood loss (orange). 
	   The ground truth data is provided by a stereo SGM\cite{hirschmuller2005accurate}-based completion of a lidar point cloud (green). }
	\label{fig_architecture}
\end{figure}

%% file: images/Architecture/ArchitecturewithImagesLatexVersion2.eps_tex
\begingroup%
  \makeatletter%
  \providecommand\color[2][]{%
    \errmessage{(Inkscape) Color is used for the text in Inkscape, but the package 'color.sty' is not loaded}%
    \renewcommand\color[2][]{}%
  }%
  \providecommand\transparent[1]{%
    \errmessage{(Inkscape) Transparency is used (non-zero) for the text in Inkscape, but the package 'transparent.sty' is not loaded}%
    \renewcommand\transparent[1]{}%
  }%
  \providecommand\rotatebox[2]{#2}%
  \newcommand*\fsize{\dimexpr\f@size pt\relax}%
  \newcommand*\lineheight[1]{\fontsize{\fsize}{#1\fsize}\selectfont}%
  \ifx\svgwidth\undefined%
    \setlength{\unitlength}{414.88660536bp}%
    \ifx\svgscale\undefined%
      \relax%
    \else%
      \setlength{\unitlength}{\unitlength * \real{\svgscale}}%
    \fi%
  \else%
    \setlength{\unitlength}{\svgwidth}%
  \fi%
  \global\let\svgwidth\undefined%
  \global\let\svgscale\undefined%
  \makeatother%
  \begin{picture}(1,0.47682935)%
    \lineheight{1}%
    \setlength\tabcolsep{0pt}%
    \put(0,0){\includegraphics[width=\unitlength]{images/Architecture/ArchitecturewithImagesLatexVersion2.eps}}%
	\scriptsize
    \put(6.53465188,0.47057112){\color[rgb]{0,0,0}\makebox(0,0)[lt]{\begin{minipage}{1.2146996\unitlength}\raggedright \end{minipage}}}%
    \put(0.08566404,0.403){\color[rgb]{0,0,0}\makebox(0,0)[lt]{\lineheight{0.78125018}\smash{\begin{tabular}[t]{l}left image\end{tabular}}}}%
    \put(0.07544017,0.22){\color[rgb]{0,0,0}\makebox(0,0)[lt]{\lineheight{0.78125018}\smash{\begin{tabular}[t]{l}right image\end{tabular}}}}%
    \put(0.05318231,0.095){\color[rgb]{0,0,0}\makebox(0,0)[lt]{\lineheight{0.78125018}\smash{\begin{tabular}[t]{l}lidar point cloud\end{tabular}}}}%
    \put(0.4,0.12){\color[rgb]{0,0,0}\makebox(0,0)[lt]{\lineheight{0.78125012}\smash{\begin{tabular}[t]{l}SGM completion\end{tabular}}}}%
    \put(0.66,0.11233634){\color[rgb]{0,0,0}\makebox(0,0)[lt]{\lineheight{0.78125012}\smash{\begin{tabular}[t]{l}depth split\end{tabular}}}}%
    \put(0.78462385,0.016){\color[rgb]{0,0,0}\makebox(0,0)[lt]{\lineheight{0.78125012}\smash{\begin{tabular}[t]{l}calibration split\end{tabular}}}}%
    \put(0.3700851,0.34){\color[rgb]{1,1,1}\makebox(0,0)[lt]{\lineheight{0.93750006}\smash{\begin{tabular}[t]{l}DepthNet\end{tabular}}}}%
    \put(0.728,0.322){\color[rgb]{1,1,1}\makebox(0,0)[lt]{\lineheight{0.93750012}\smash{\begin{tabular}[t]{l}CalibNet\end{tabular}}}}%
    \put(0.32875371,0.44339434){\color[rgb]{0,0,0}\makebox(0,0)[lt]{\lineheight{0.78125006}\smash{\begin{tabular}[t]{l}ProbDepthNet\end{tabular}}}}%
    \put(0.908,0.1815){\color[rgb]{0,0,0}\makebox(0,0)[lt]{\lineheight{0.78125}\smash{\begin{tabular}[t]{l}loss $\mathcal{L}$\end{tabular}}}}%
    \put(0.615,0.185){\color[rgb]{0,0,0}\makebox(0,0)[lt]{\lineheight{0.78125018}\smash{\begin{tabular}[t]{l}loss $\mathcal{L}$\end{tabular}}}}%
    \put(0.89,0.3994627){\color[rgb]{0,0,0}\makebox(0,0)[lt]{\lineheight{0.78125006}\smash{\begin{tabular}[t]{l}$\mu_0$...$\mu_K$ \end{tabular}}}}%
    \put(0.89,0.35216095){\color[rgb]{0,0,0}\makebox(0,0)[lt]{\lineheight{0.78125006}\smash{\begin{tabular}[t]{l}$\tilde{s}_0$...$\tilde{s}_K$\end{tabular}}}}%
    \put(0.89,0.30295983){\color[rgb]{0,0,0}\makebox(0,0)[lt]{\lineheight{0.78125006}\smash{\begin{tabular}[t]{l}$\tilde{\lambda}_0$...$\tilde{\lambda}_K$\end{tabular}}}}%
    \put(0.57,0.39752252){\color[rgb]{0,0,0}\makebox(0,0)[lt]{\lineheight{0.78125006}\smash{\begin{tabular}[t]{l}$\mu_0$...$\mu_K$ \end{tabular}}}}%
    \put(0.57,0.35022076){\color[rgb]{0,0,0}\makebox(0,0)[lt]{\lineheight{0.78125006}\smash{\begin{tabular}[t]{l}$s_0$...$s_K$\end{tabular}}}}%
    \put(0.57,0.30101969){\color[rgb]{0,0,0}\makebox(0,0)[lt]{\lineheight{0.78125006}\smash{\begin{tabular}[t]{l}$\lambda_0$...$\lambda_K$\end{tabular}}}}%
  \end{picture}%
\endgroup%

%% file: text/Method/MonoSceneFlow.tex
\subsection{Monocular Scene Flow}
\label{sec_monosf}
This section presents the Mono-SF optimization framework, structured as follows: 
First, the decomposition of the scene into piecewise planar surface elements and rigid bodies is described.
Second, the optimization is formulated as an energy minimization problem combining
 a) multi-view geometry-based photometric distance,
 b) the probabilistic single-view depth estimates of ProbDepthNet and
 c) scene model smoothness priors.
Finally, the inference and initialization of the optimization problem are presented.

\textbf{Monocular Scene Flow Model:} 
Following previous object scene flow approaches
   \cite{Menze2015CVPR, behl2017bounding, MENZE201860},
   the main assumption is that, in particular, a traffic scene can be approximated
 by a set of piecewise planar surface elements to represent the structure of the scene
 and a set of rigid bodies to represent the motion (see Fig. \ref{fig_variables}).  
Formally, the reference image is divided into a set of superpixels each one representing a 3D plane.
 Each 3D plane is defined by its normal $\vn_i \in \mathbb{R}^3$,
 scaled by the inverse distance of the plane to the camera 
 to encode the 3D position $\mX$ of each point on the plane by $\vn_i^T \mX = 1$.
The set of rigid bodies consists of the background 
 as well as other traffic participants such as pedestrians or vehicles
 detected by an instance segmentation. 
Even though a pedestrian does not undergo a rigid body motion,
 at a certain scale, it can be approximated by its dominant rigid body transformation as motivated by \cite{MENZE201860}.
Each rigid body is represented by its 6D motion $\mT_{j} \in SE(3)$.
Additionally, each superpixel is associated with one rigid body and with the pixels $\mathcal{R}_i$
 of the corresponding superpixel.
\input{images/SceneModel/Figure.tex}

\textbf{Energy Minimization Problem:} 
The main idea of Mono-SF is that the scene geometry and motion
should be consistent in terms of warping the reference image $I_0$ in the consecutive image $I_1$
and consistent to the depth distributions $p(d \mid I_0)$ and $p(d \mid I_1)$ provided by ProbDepthNet.
Formally, Mono-SF jointly optimizes the 6D motion of each rigid body $\mT_{j}$ and 3D normal of each plane $\vn_i$
 as an energy minimization problem. 
The energy term $E$ consists of unary data terms $\Phi(\vp_0,\vn_i,\mT_{j})$ for each pixel $\vp_{0}$ 
and pairwise smoothness terms $\Psi(\vn_i, \vn_j)$ for each two planes $\vn_k$ and $\vn_l$ adjacent in the image $k,l \in \mathcal{N}$:
\begin{equation}
    E = \sum_{\vn_i} \sum_{\vp_{0} \in \mathcal{R}_i} \Phi(\vp_{0},\vn_i,\mT_{j}) 
    + \sum_{k,l \in \mathcal{N} } \Psi(\vn_k, \vn_l)
    \label{eqn_energy}
\end{equation}
$\mT_{j}$ is the rigid body corresponding to the plane $\vn_i$.

The unary terms $\Phi(\vp_{0},\vn_i,\mT_{j})$ consist of two parts.
First, $\Phi^{pho}(\vp_{0},\vn_i,\mT_{j})$ minimizes an appearance-based photometric distance 
between pixel $\vp_0$ and its projected position in the consecutive image. 
Second, $\Phi^{svd}_{t}(\vp_{0},\vn_i,\mT_{j})$ prefers a 3D position consistent to the estimated depth probabilities of ProbDepthNet 
at time $t=0$ and $t=1$:
\begin{eqnarray}
\Phi(\vp_{0},\vn_i,\mT_{j}) =&  \Theta_{0} & \Phi^{pho}(\vp_{0},\vn_i,\mT_{j}) \nonumber \\
 + & \Theta_{1} & \sum_{t \in \{0,1\}} \Phi^{svd}_{t}(\vp_{0},\vn_i,\mT_{j})
\end{eqnarray}
The terms are weighted by $\Theta_{0}$ or $\Theta_{1}$, respectively. 
The photometric distance $\Phi^{pho}(\vp_0,\vn_i,\mT_{j})$ rates the similarity 
of the two corresponding image positions $\vp_{0}$ and $\vp_{1}$ as the hamming distance of their respective $5 \times 5$ Census descriptors 
   \cite{zabih1994non} truncated at $\tau_0$. 
The corresponding image coordinates $\vp_1$ in the second image $I_{1}$ are defined by a homography \cite{hartley2003multiple} considering
the 3D normal $\vn_i$ and the motion of the corresponding rigid body $\mT_{j}$:
\begin{equation}
\vp_{1} = \mK(\mR_j - \vt_j \vn_i^T) \mK^{-1} \vp_0
\label{eqn_p1}
\end{equation}
$\mR_j$ and $\vt_j$ is the decomposition of $\mT_{j}$ into rotation matrix and translation vector.
$\mK$ is the intrinsic camera matrix.

The term $\Phi_{t}^{svd}(\vp_0,\vn_i,\mT_{j})$ rates the consistency of the depth of pixel $\vp_0$ based
 on the ProbDepthNet estimates. 
Whereas the depth $d_{0}(\vp_0,\vn_i)$ at time $t=0$ is directly defined by the corresponding scaled normal vector $\vn_i$,
 the motion of the corresponding rigid body $\mT_{j}$
 needs to be considered to derive the depth $d_{1}(\vp_0,\vn_i,\mT_{j})$ at time $t=1$. 
Both depth values are rated by the negative log-likelihood of the probability
 provided by ProbDepthNet for their respective image $I_t$ and image coordinate $\vp_t$:  
\begin{equation}
\Phi^{svd}_{t}(\vp_0,\vn_i,\mT_{j}) =- \log  p_{\vp_{t}} \left( d_{t}(\vp_{0},\vn_i,\mT_{j}) \mid I_{t} \right)
\label{eqn_sid}
\end{equation}
The image coordinates $\vp_1$ are again defined as in Eq. \eqref{eqn_p1}.

The previous data terms include the single-view depth information and multi-view geometry-based photometric distance.
Additionally, scene model priors are integrated similar to \cite{Menze2015CVPR} as
 pairwise smoothness terms $\Psi(\vn_k, \vn_l)$ preferring a smooth structure in terms of depth $\Psi^{d}(\vn_k, \vn_l)$
   and orientation $\Psi^{ori}(\vn_k, \vn_l)$, each part weighted by $\Theta_{2}$ or $\Theta_{3}$:
\begin{equation}
\Psi(\vn_k, \vn_l) = \Theta_{2} \Psi^{d}(\vn_k, \vn_l) + \Theta_{3} \Psi^{ori}(\vn_k, \vn_l)
\end{equation}
For each shared boundary pixel $\vp_0 \in \mathcal{B}_{k,l}$ of plane $\vn_k$ and $\vn_l$,
 a difference in depth is penalized:
\begin{equation}
\Psi^{d}(\vn_k, \vn_l) = \sum_{\vp_0 \in \mathcal{B}_{k,l}} \min \left(| d_0(\vp_0,\vn_k) - d_0(\vp_0,\vn_l) |, \tau_1 \right)
\end{equation}
Analogously, a smooth orientation of planes adjacent in the image is preferred
by measuring the similarity of the normal vectors $\vn_k$ and $\vn_l$:  
\begin{equation}
\Psi^{ori}(\vn_k, \vn_l) = \min \left(1 - \frac{ | \vn_k \vn_l |}{ ||\vn_k|| ||\vn_l||} , \tau_2 \right)
\end{equation}
Both smoothness terms are truncated by $\tau_1$ or $\tau_2$ to regard discontinuities in the depth or orientation,
 for example between different objects. 
The hyper-parameters $\Theta$ and $\tau$ are defined differently according to the rigid body type, background or object,
 and differently for adjacent planes belonging to different rigid bodies. 
These dependencies are neglected in the previous equations for ease of reading. 

\textbf{Inference:}
The scene flow estimation is formulated as the energy minimization problem in Eq. \eqref{eqn_energy}. 
Assuming a suitable initialization, that will be discussed in the next section,
 an iterative optimization approach can be applied. 
Following the proposed optimization of the object scene flow methods 
   \cite{Menze2015CVPR, behl2017bounding}, particle max-product belief propagation is used
 for 10 iterations with 5 particles for each 6D rigid body motion and 10 particles for each 3D normal vector. 

\textbf{Initialization:}
The optimization problem needs a suitable initialization of all variables. 
In the first step, the set of rigid bodies is initialized including their scale-aware 6D motions. 
Traditionally, the known camera height or an additional inertial measurement unit is used for scale-aware monocular visual odometry in the automotive domain.
However, this only provides scale information for the camera ego-motion.
The key idea applied here is to integrate single-view depth information to provide the metric scale. 
In contrast to \cite{barnes2018driven,yin2017scale,yang2018deep}, we apply this idea additionally for scale-aware pose estimation of moving objects.
First, object instances in the images $I_0$ and $I_1$ detected by a Mask R-CNN 
   \cite{he2017mask}
 (implementation of \cite{mxmaskrcnn})
  are paired based on sparse flow correspondences $(\vp^i_0, \vp^i_1)$ 
   \cite{geiger2011stereoscan} 
 using a simple voting scheme.
Each object instance, as well as the background, builds a rigid body.
Second, the 6D motion $\mT_{j} \in SE(3)$ of each rigid body is optimized jointly with
 a set of 3D points $\mX_i \in \mathcal{X}$ (one for each flow correspondence lying in the corresponding instance masks)
 by minimizing
\begin{equation}
    \sum_{\mX_i \in \mathcal{X}} \sum_{t \in \{0,1\}} \Theta_{4} \Phi_t^{proj}(\vp^i_t, \mX_i, \mT_j) + \Phi_t^{svd}(\mX_i, \mT_j) .
	\label{eqn_init}
\end{equation}
$\Phi_t^{proj}(\vp^i_t, \mX_i, \mT_j)$ is the reprojection error of $\mX_i$ with respect to the flow-based image positions $\vp^i_t$ weighted by $\Theta_{4}$. 
$\Phi_t^{svd}(\mX_i, \mT_j)$ rates the consistency of the 3D points $\mX_i$
 to the ProbDepthNet estimates analogously to Eq. \eqref{eqn_sid}.
 The energy term of Eq. \eqref{eqn_init} is optimized using the Levenberg-Marquardt solver implemented in
   \cite{Kuemmerle2011}. 

Subsequently, the set of 3D planes is initialized. 
First, a dense depth map is computed based on a semi-global matching adapted to the monocular case similarly to
   \cite{yamaguchi2013robust, bai2016exploiting}. 
Again, the depth estimates are additionally rated by the ProbDepthNet estimates. 
Second, the superpixels including their 3D normal $\vn_i$ are initialized using the approach in \cite{sps}. 
The pixels of a plane are enforced to be of the same instance to get a unique association with a rigid body.
%

%% file: images/SceneModel/Figure.tex
\begin{figure}[t]
\centering
	\setlength\tabcolsep{1 pt}
	\small 
\begin{tabular}{c c}
6D motions $\mT_j$ of rigid bodies& 3D normals $\vn_i$ of planes \\
\includegraphics[width=0.23\textwidth]{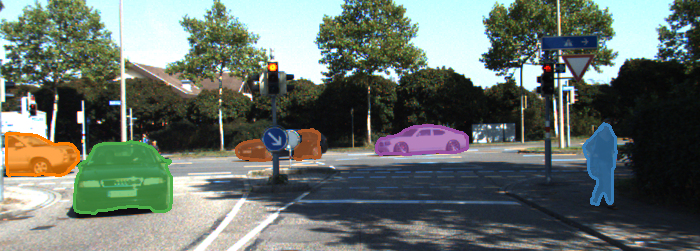} &
\includegraphics[width=0.23\textwidth]{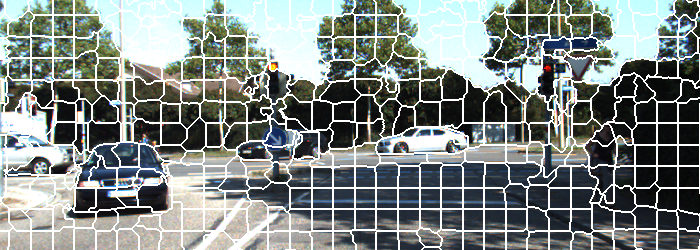}
\end{tabular}
  \caption{Variables of Mono-SF model and energy minimization problem are the 6D rigid body motions $\mT_j$ of moving objects (colored in the left image) and the background
  as well as the 3D scaled normals $\vn_i$ of superpixel planes (boundaries in the right image).}
  
	\label{fig_variables}
\end{figure}

%% file: text/Experiments/Experiments.tex
\input{images/ExampleProbDepth/Figure.tex} 
\input{images/GeneralizationProbDepthNet/Figure.tex}
In the first part of this section ProbDepthNet is analyzed:
Qualitative results of ProbDepthNet are shown, 
the generalization capabilities to other datasets are presented
and an ablation study confirms the importance of the recalibration technique
to provide well-calibrated distributions.
In the second part, the Mono-SF optimization framework is evaluated
by showing qualitative results and 
a quantitative evaluation with respect to several state-of-the-art methods.
Additionally, two ablation studies confirm the claimed ProbDepthNet design for Mono-SF 
and support the importance of the individual components of Mono-SF.
\subsection{Probabilistic Single-View Depth Estimation}

The experiments are conducted on a ProbDepthNet model trained for the 
 KITTI scene flow training set \cite{MENZE201860}.
The model is trained on 33 sequences of the KITTI raw dataset
 that are not part of the scene flow set. 
Around 75\% / 25\% of the sequences are used for training DepthNet / CalibNet. 
It is trained for 15 epochs using Adam optimizer 
    \cite{kingma2014adam} with a learning rate of $10^{-4}$ halved every 5 epochs and a small batch size of 4. 
The input images are scaled to a size of $512 \times 256$ and a mixture of Gaussians with 8 components is used. 

Fig. \ref{fig_exampleProbDepth} shows exemplary the output of ProbDepthNet. 
The variances visually correlate with challenging parts of the scene such as object boundaries or poles. 
The estimated recalibrated variances $\tilde{s}_0$ provided by CalibNet
 are significantly higher than the variances $s_0$. 
The generalization capabilities of ProbDepthNet trained for KITTI are visualized 
  by the qualitative results on the Make3D \cite{saxenamake3d}
  and Cityscapes \cite{cordts2016cityscapes}
  dataset in Fig. \ref{fig_generalizationProbDepthNet}. 
Please see the supplementary material for more qualitative results and discussions.
\input{images/Recalibration/Figure.tex}

The following ablation study analyzes the proposed recalibration by adding the CalibNet trained on a hold-out split.
Our proposed training by minimizing the negative log-likelihood (NLL) is related to the approach in \cite{kendall2017uncertainties}.
But, to provide a comparison of different probabilistic approaches, 
 the DepthNet part is also trained using a multi-hypothesis strategy ('Hypo \cite{ICKMB18}') similar to \cite{ICKMB18}
 or transformed to its 'assumed density filtering'-counterpart ('ADF\cite{gast2018lightweight}') as proposed by \cite{gast2018lightweight}.
Fig. \ref{fig_recalibration} shows the mean NLL on the KITTI scene flow set (which is not part of the training data) every 1000 training steps. 
 In the bottom plot of Fig. \ref{fig_recalibration}, the calibration of the final models is evaluated. 
The frequency of ground truth depth values inside a given interval
 should be the same as the cumulative probability of the estimated distribution. 
The impact of overfitting effects varies among the different approaches -- 
 but all approaches suffer from such an effect and provide overconfident estimates.
Furthermore, CalibNet is validated as an useful recalibration technique applicable to different probabilistic approaches.

For integration in Mono-SF, a model is additionally pre-trained on Cityscapes \cite{cordts2016cityscapes}.
Compared to previous non-probabilistic methods for single-view depth estimation such as \cite{godard2016deepdepth,kuznietsov2017semi,fu2018deep},
 the main benefit of ProbDepthNet is providing well-calibrated depth distributions.
However, in addition to correct uncertainties, the underlying estimates should have sufficient quality as well.
A quantitative evaluation (see supplementary material) shows that the accuracy of the depth estimates 
 represented by the total means of the distributions is comparative to \cite{godard2016deepdepth,kuznietsov2017semi} and slightly below \cite{fu2018deep}.

\subsection{Monocular Scene Flow}
\input{images/Results/Figure.tex}
\input{images/ResultsCityscapes/Figure.tex}
\input{tables/Experiments/KittiSceneFlowTrain.tex}

Mono-SF estimates the 3D scene flow from monocular images focusing on dynamic traffic scenes,
 which means providing the 3D position and 3D motion of each pixel.
The following results and evaluations are based on the equivalent representation 
 as the depth of each pixel at both times ($t=0$, $t=1$) and the optical flow. 
Thereby, the 3D position and the ability of the approaches to predict a 3D point from $t=0$ to $t=1$ based in its 3D motion is evaluated.
Exemplary qualitative results of Mono-SF are shown for the KITTI 
  \cite{MENZE201860} (see Fig. \ref{fig_exampleResults}) and Cityscapes dataset \cite{cordts2016cityscapes}  (see Fig. \ref{fig_exampleResultsCS}).
Please see the supplementary material for further results. 

The quantitative evaluation is based on the KITTI scene flow dataset
	\cite{MENZE201860}, 
 which reports the frequencies of errors for the depth at time $t=0$ (D1) and $t=1$ (D2) and the optical flow (Fl). 
An estimate is considered as an error if it exceeds a threshold of 3 pixels
 and 5\% in terms of stereo disparity or optical flow endpoint error. 
Furthermore, an estimate is only defined as a valid scene flow estimate (SF) if it fulfills all the D1, D2, and Fl metrics. 
All metrics are evaluated separately for moving objects (fg), the static scene (bg) and both combined (all). 
 
We propose four categories of state-of-the-art monocular baseline methods.
In the first category are the multi-task networks, GeoNet 
    \cite{yin2018geonet},
DF-Net 
    \cite{zou2018df}
and EveryPixel 
    \cite{yang2018every}.
These CNNs are trained in an unsupervised manner and are able to provide single-view depth estimates for both images and optical flow estimates. 
For the GeoNet and DF-Net, their published code and models are used. 
The results of the EveryPixel approach are stated in their paper 
    \cite{yang2018every} 
(D2 metric is excluded as it seems to be inconsistent). 
As a second category, single-view depth estimation ('LRC\cite{godard2016deepdepth}' or 'DORN \cite{fu2018deep}') and optical flow estimation ('MirrorFlow \cite{hur2017mirrorflow}' or 'HD$^3$-F \cite{yin2018hierarchical}') are combined as individual tasks. 
Due to the fact that the published models of 'DORN \cite{fu2018deep}' and 'HD$^3$-F \cite{yin2018hierarchical}' used parts of the dataset for training,
 these methods are disregarded for ranking.
The third group comprises the multi-body or non-rigid SfM-based methods
DMDE \cite{ranftl2016dense} and S.Soup \cite{kumar2017monocular}.
The fourth category consists of the methods MFA \cite{kumar2019motion}, Mono-Stixel \cite{monostixel2} and our Mono-SF approach,
which are methods that fuse single-view depth information with multi-view geometry.
DMDE, S.Soup, and MFA were only evaluated on its depth estimates capped at 50m using a mean absolute relative error (MRE). 
For the Mono-Stixel approach, the authors provide us the results on a scene flow metric using MirrorFlow 
  \cite{hur2017mirrorflow}
 and LRC 
  \cite{godard2016deepdepth} 
 as inputs.
The results of the quantitative evaluation are shown in Table \ref{tab_evaluationKittiTrain}.
To the best of our knowledge, it is the first time that these methods are evaluated and compared as a scene flow estimation problem. 
The results show that the methods of the fourth group that combine single-view depth and multi-view geometry outperforms the other methods. 
Mono-SF shows the best rating on most of the metrics and especially outperforms previous methods on the scene flow (SF) metrics.
The approach and implementation of Mono-SF is currently not focused on runtime and needs around 41 seconds per image on a single CPU-core.
Mono-SF was also submitted to the KITTI scene flow benchmark (see Table \ref{tab_resultsKittiTest}).
Mono-SF is the first monocular method and would have been ranked at the 13th place with respect to the 21 published stereo scene flow methods.
\input{tables/Experiments/KittiSceneFlowTest.tex}
\subsection{Ablation Studies}
\label{seq_ablationStudies}
\input{tables/Experiments/AblationProbDepthNet.tex}
\input{tables/Experiments/AblationMonoSF.tex}
To analyze the importance of the proposed ProbDepthNet design, 
the results of four Mono-SF variants based on different single-view depth estimations are provided in Table \ref{tab_ablationProbDepthNet}.
The two Mono-SF variants "Mono-SF (LRC \cite{godard2016deepdepth})" and "Mono-SF (w/o prob. depth)" 
 utilized CNNs that provide only single-view depth values instead of depth distributions.
Whereas "Mono-SF (LRC \cite{godard2016deepdepth})" is based on the LRC method for single-view depth estimation,
"Mono-SF (w/o prob. depth)" is based on the non-probabilistic estimates of ProbDepthNet represented by the total means of the distributions.
The depth values are integrated by assuming the same Gaussian distribution (determined on a test set) for all pixels.
Mono-SF based on the probabilistic ProbDepthNet ("Mono-SF") outperforms both.
This supports the claimed ProbDepthNet design to provide single-view depth estimates in a probabilistic form. 
Furthermore, the improvements compared to a variant based on the ProbDepthNet excluding CalibNet "Mono-SF (w/o recalib.)"
 support that the recalibration technique is an essential component.
 
In Table \ref{tab_ablationMonoSF}, the individual components of the Mono-SF optimization framework are analyzed
by removing some parts of the proposed energy minimization problem (setting their weights to zero).
The initialization of Mono-SF described in Sec. \ref{sec_monosf} is denoted by the row without checkmarks. 
Compared to this initialization, the scene flow formulation of Mono-SF results in further improvement.
Additionally, the ablation study shows that each part of the energy term contributes to the final performance;
the multi-view geometry, the single-view depth information and the scene model smoothness priors.

%% file: images/ExampleProbDepth/Figure.tex
\begin{figure*}[bth]
\centering
\setlength\tabcolsep{1 pt}
\begin{tabular}{c c c c c}
\small Image & \small Ground truth depth & \small Mean depth $\mu_0$ & \small Variance $s_0$ & \small Recalib. variance $\tilde{s}_0$ \\
\includegraphics[width=0.195\textwidth]{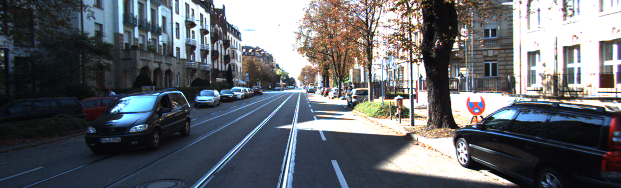} &
\includegraphics[width=0.195\textwidth]{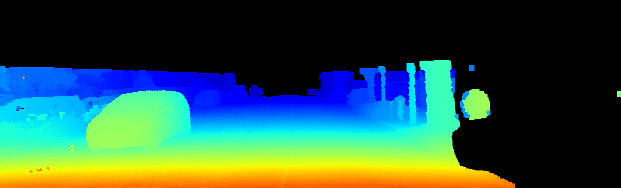} &
\includegraphics[width=0.195\textwidth]{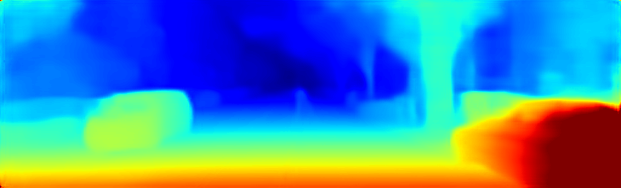} &
 \includegraphics[width=0.195\textwidth]{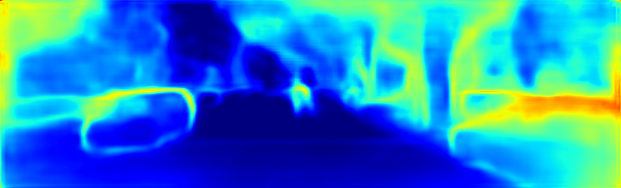} &
\includegraphics[width=0.195\textwidth]{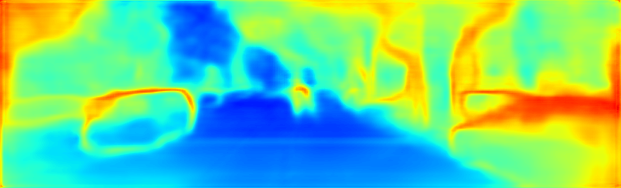} \\
\includegraphics[width=0.195\textwidth]{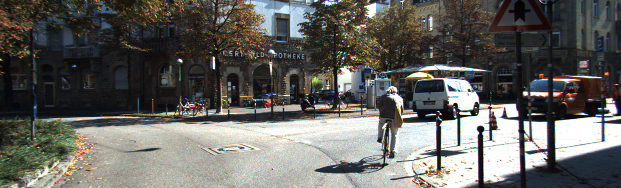} &
\includegraphics[width=0.195\textwidth]{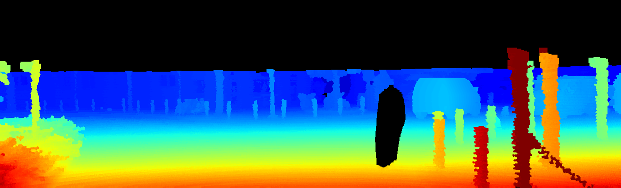} &
\includegraphics[width=0.195\textwidth]{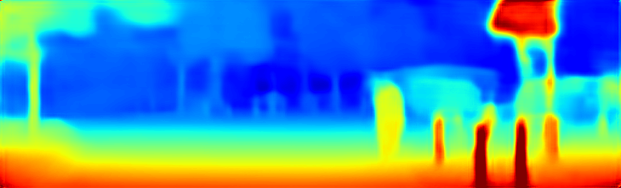} &
 \includegraphics[width=0.195\textwidth]{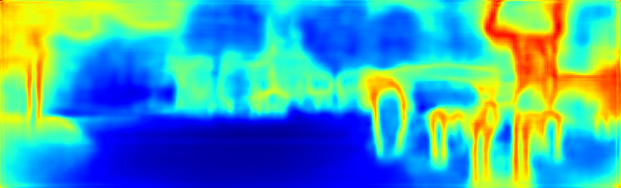} &
\includegraphics[width=0.195\textwidth]{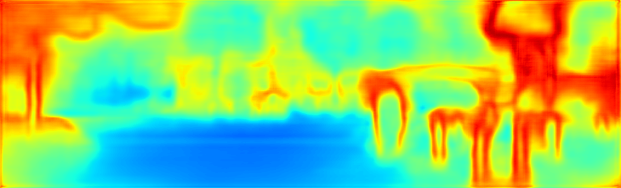} 
\end{tabular}
\caption{Exemplary estimates of ProbDepthNet on KITTI scene flow set \cite{Menze2015CVPR} 
  for the first component of the mixture of Gaussians excluding the weight. 
  The color encodes the inverse depth from close (red) to far (blue) 
  or high variance (red) to low variance (blue).}
\label{fig_exampleProbDepth}
\end{figure*}

%% file: images/GeneralizationProbDepthNet/Figure.tex
\begin{figure}[tbh]
\centering
\setlength\tabcolsep{1 pt}
\begin{tabular}{c c c}
\includegraphics[width=0.153\textwidth]{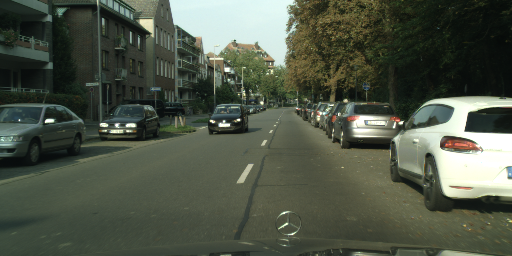} & 
\includegraphics[width=0.153\textwidth]{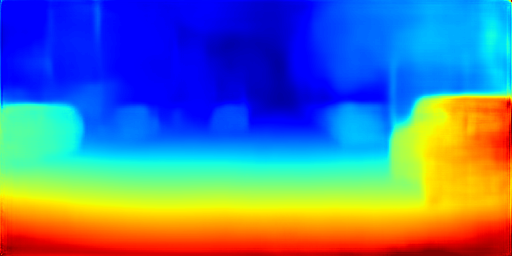} & 
\includegraphics[width=0.153\textwidth]{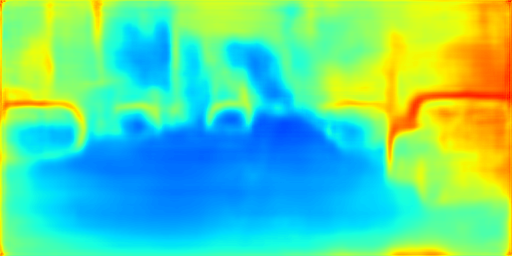} \\
\includegraphics[width=0.153\textwidth]{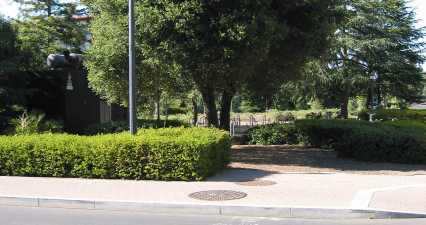} &
\includegraphics[width=0.153\textwidth]{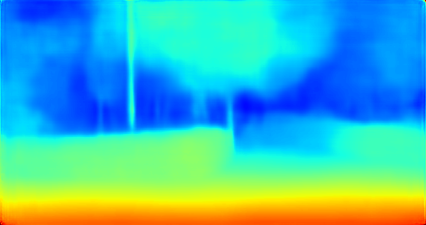} & 
\includegraphics[width=0.153\textwidth]{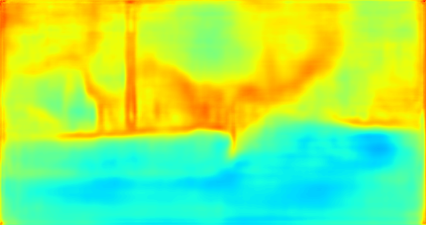}
\end{tabular}
\caption{Generalization of ProbDepthNet (trained on KITTI) on Cityscapes \cite{cordts2016cityscapes} (top) and central crop of Make3D \cite{saxenamake3d} (bottom).
The figure shows the estimates based on the left image in the form of
the mean depth values $\mu_0$ (middle) and recalibrated log-variances $\tilde{s}_0$ (right)
of the first component. }
\label{fig_generalizationProbDepthNet}
\end{figure}

%% file: images/Recalibration/Figure.tex
\begin{figure}[b]
\begin{picture}(100,100)
\put(0,0){\includegraphics[width=0.47\textwidth]{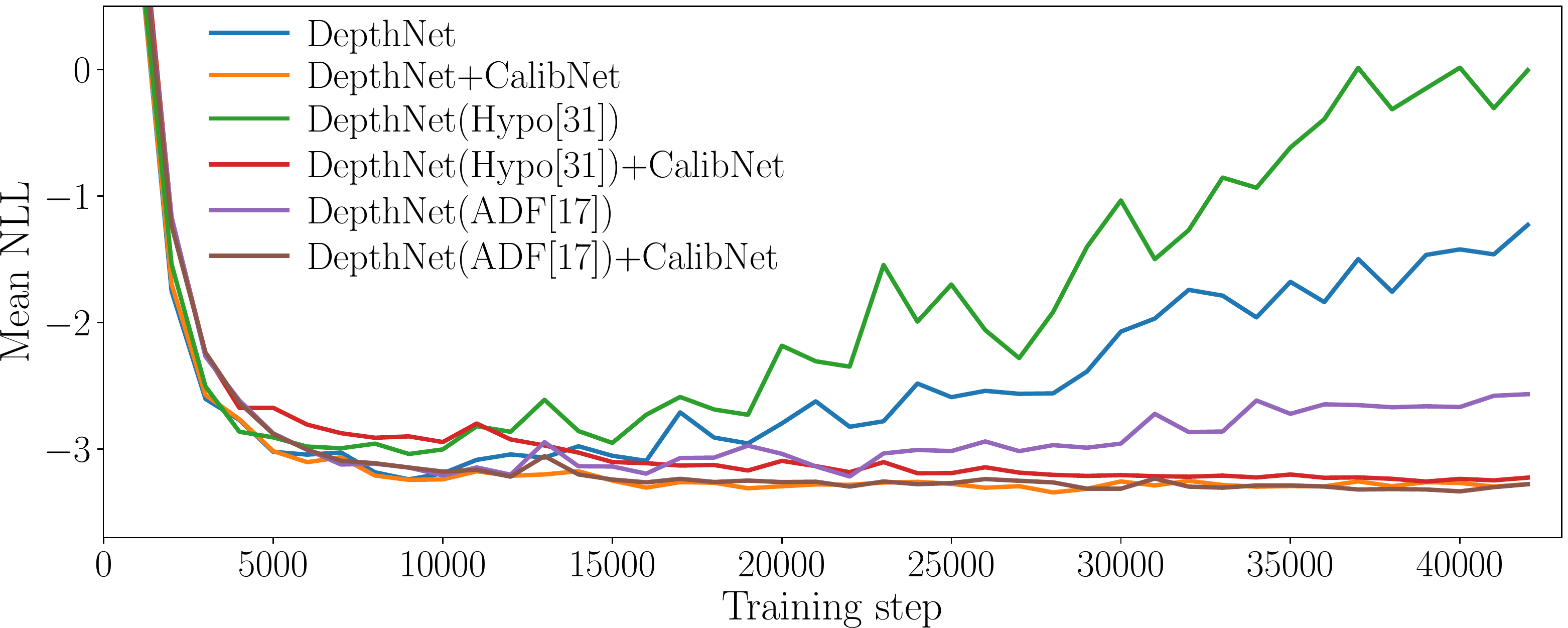}}
\put(192,19){\includegraphics[width=0.0205\textwidth]{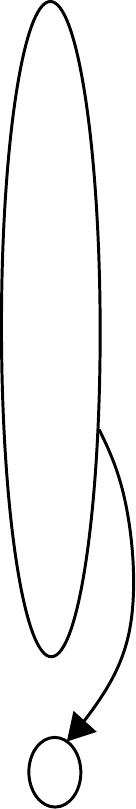}}
\put(135,55){\includegraphics[width=0.1\textwidth]{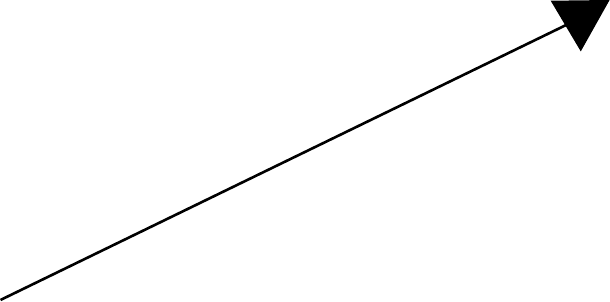}}
\put(202,25.5){\scriptsize +CalibNet}
\put(140,62){\rotatebox{25.1}{\scriptsize overfitting}}
\end{picture} \\ [-3pt]
\begin{picture}(100,100)
\put(0,0){\includegraphics[width=0.47\textwidth]{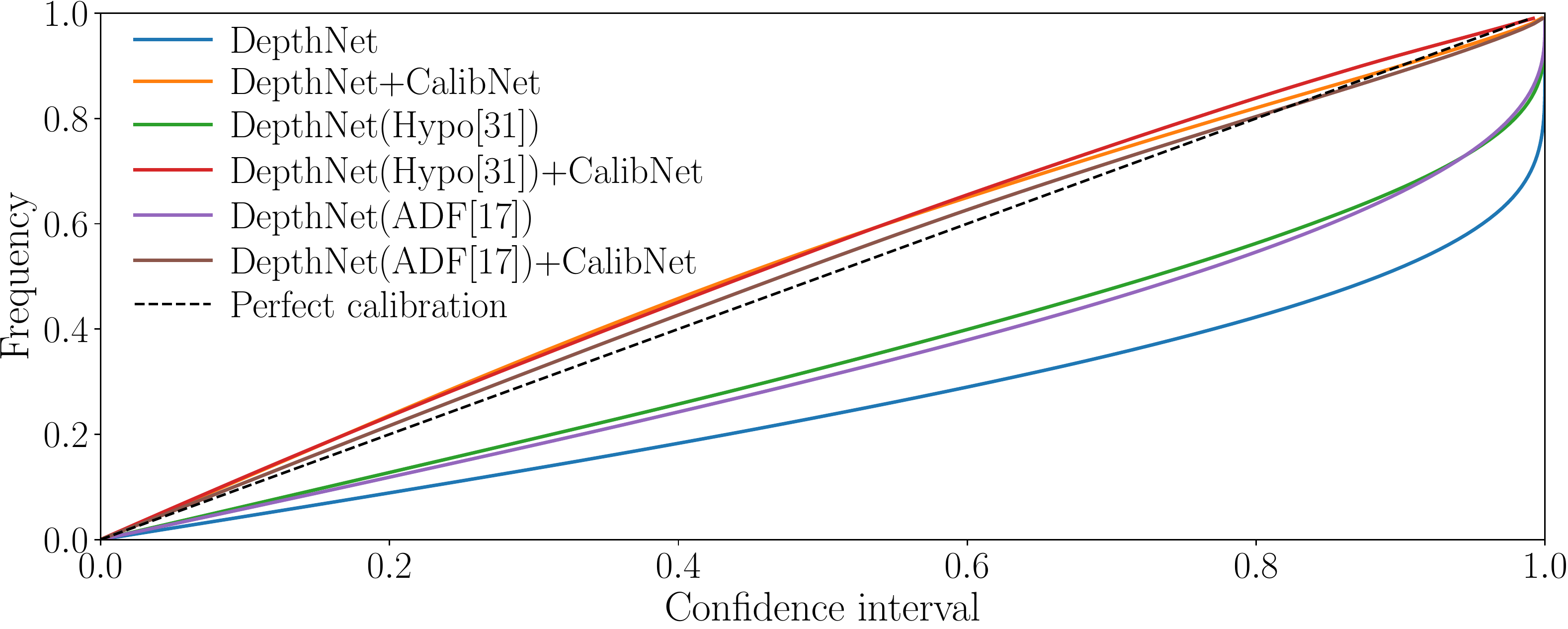}}
\put(130,30){\includegraphics[width=0.01\textwidth]{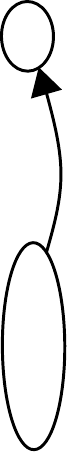}}
\put(135,51){\scriptsize +CalibNet}
\end{picture}

\caption{Top: Mean negative log-likelihood (NLL) of ProbDepthNet on the KITTI scene flow set over the training process; 
  Bottom: Calibration plot comparing the frequency of ground truth depth values lying in a given confidence interval.
  This frequency is equal to the confidence interval for a perfect calibrated model (dotted line).   
  By including CalibNet for recalibration the overfitting effect is compensated and a better calibrated
  model achieved. }
\label{fig_recalibration}
\end{figure}
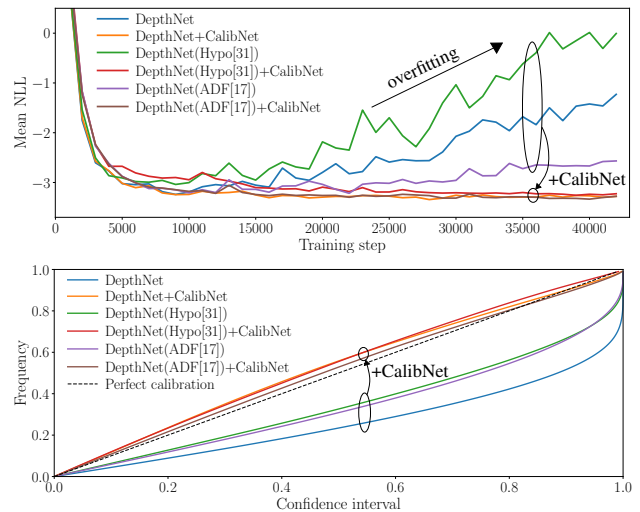

%% file: images/Results/Figure.tex
\begin{figure*}[bth]
\centering
\setlength\tabcolsep{1 pt}
\begin{tabular}{c c c c c c c}
&\small{Images} & \small{Ground truth} & \small{MirrorFlow+LRC\cite{hur2017mirrorflow,godard2016deepdepth}} & \small{Mono-Stixels \cite{monostixel2}} & \small{Mono-SF} \\
\raisebox{-.45\normalbaselineskip}[0pt][0pt]{\rotatebox[origin=c]{90}{\small Scenario 1}}&
\includegraphics[width=0.188\textwidth]{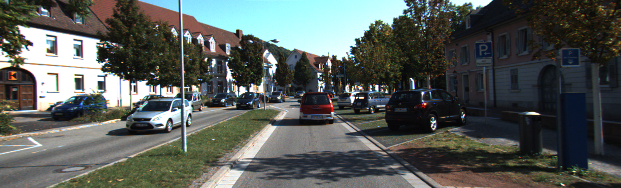} &
\includegraphics[width=0.188\textwidth]{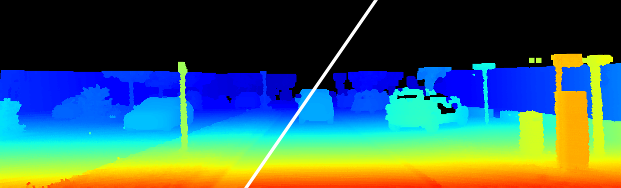} &
\includegraphics[width=0.188\textwidth]{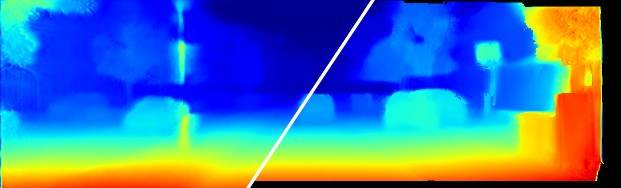} &
\includegraphics[width=0.188\textwidth]{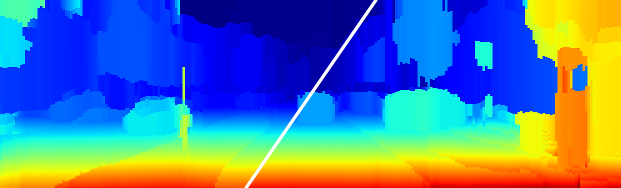} &
\includegraphics[width=0.188\textwidth]{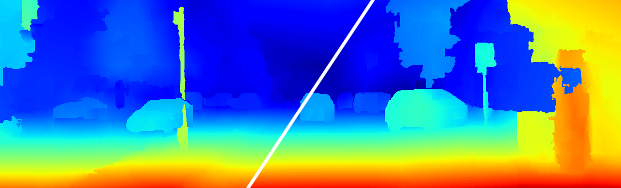} & 
\raisebox{\normalbaselineskip}[0pt][0pt]{\rotatebox[origin=c]{-90}{\small Depth}}
\vspace{-0.7mm}\\
&
\includegraphics[width=0.188\textwidth]{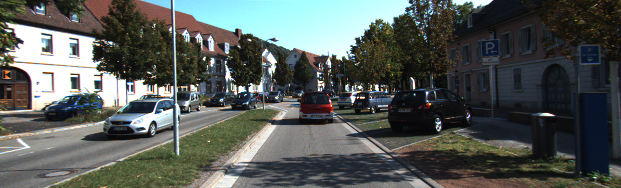} &
\includegraphics[width=0.188\textwidth]{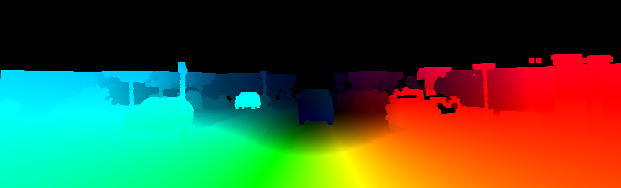} &
\includegraphics[width=0.188\textwidth]{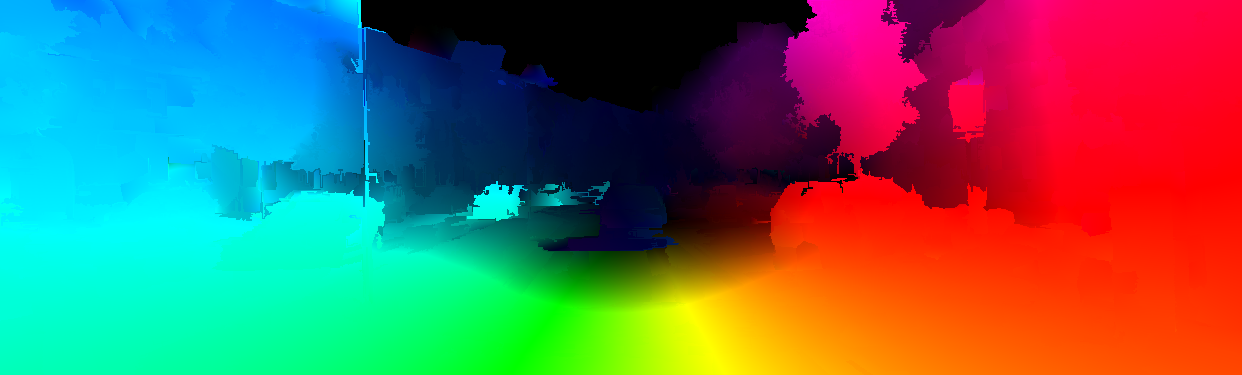} &
\includegraphics[width=0.188\textwidth]{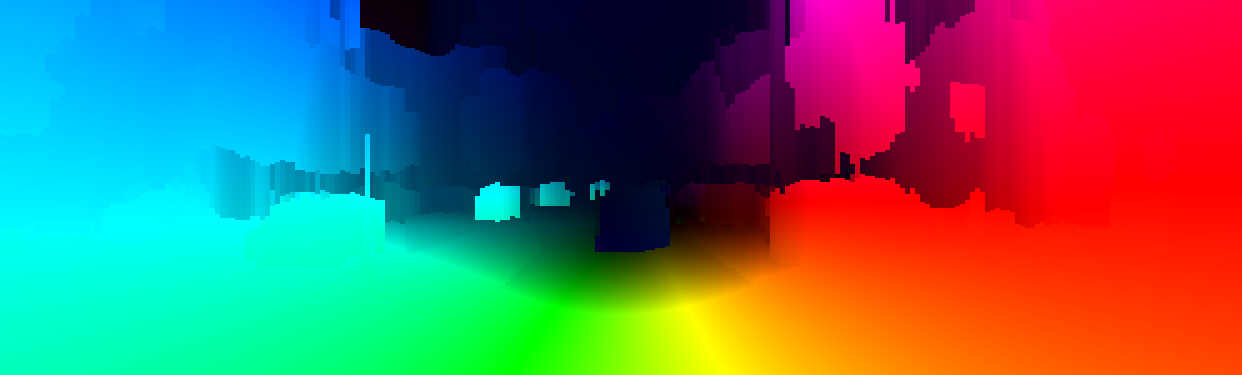} &
\includegraphics[width=0.188\textwidth]{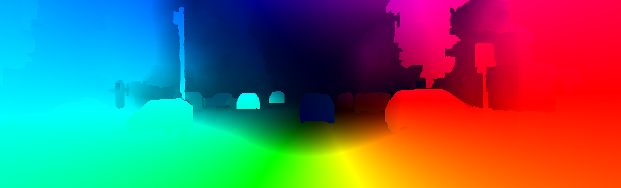} &  
\raisebox{\normalbaselineskip}[0pt][0pt]{\rotatebox[origin=c]{-90}{\small Flow}}
\vspace{-0.2mm} \\ \hline \vspace{-3mm} \\ 
\raisebox{-.45\normalbaselineskip}[0pt][0pt]{\rotatebox[]{90}{\small Scenario 2}}&
\includegraphics[width=0.188\textwidth]{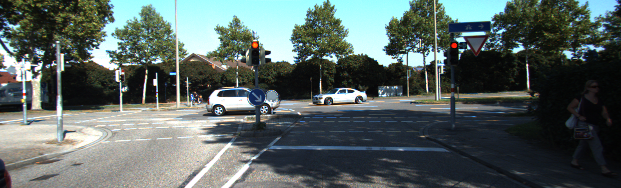} &
\includegraphics[width=0.188\textwidth]{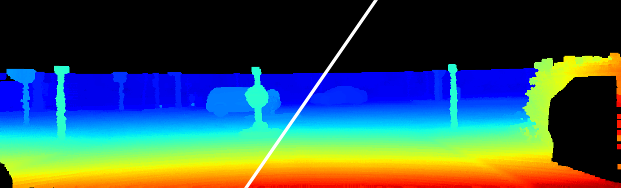} &
\includegraphics[width=0.188\textwidth]{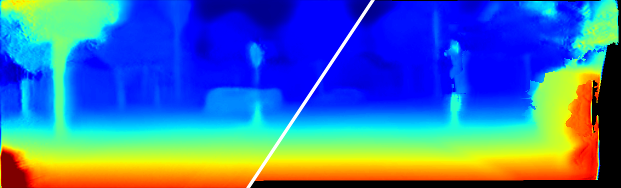} &
\includegraphics[width=0.188\textwidth]{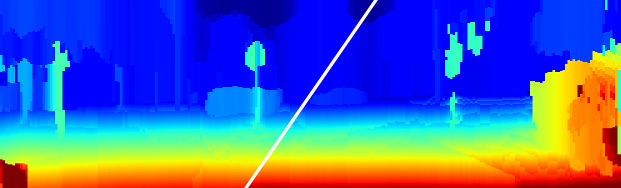} &
\includegraphics[width=0.188\textwidth]{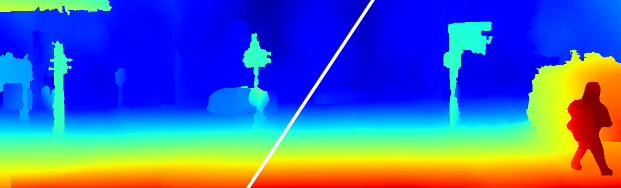} & 
\raisebox{\normalbaselineskip}[0pt][0pt]{\rotatebox[origin=c]{-90}{\small Depth}}
\vspace{-0.7mm}\\
&
\includegraphics[width=0.188\textwidth]{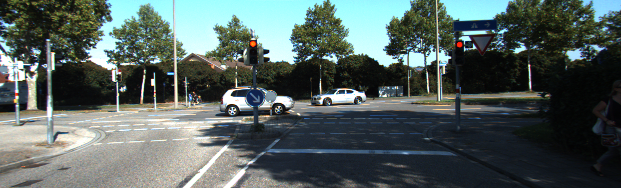} &
\includegraphics[width=0.188\textwidth]{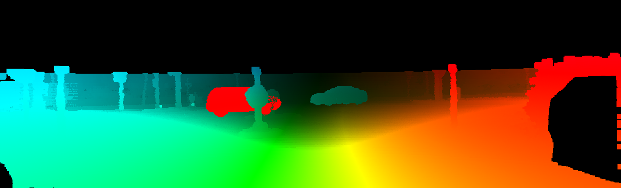} &
\includegraphics[width=0.188\textwidth]{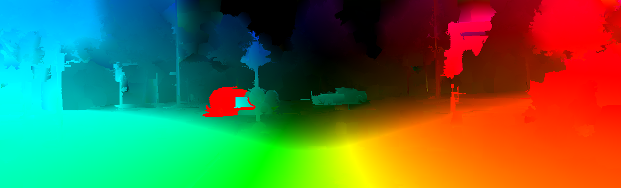} &
\includegraphics[width=0.188\textwidth]{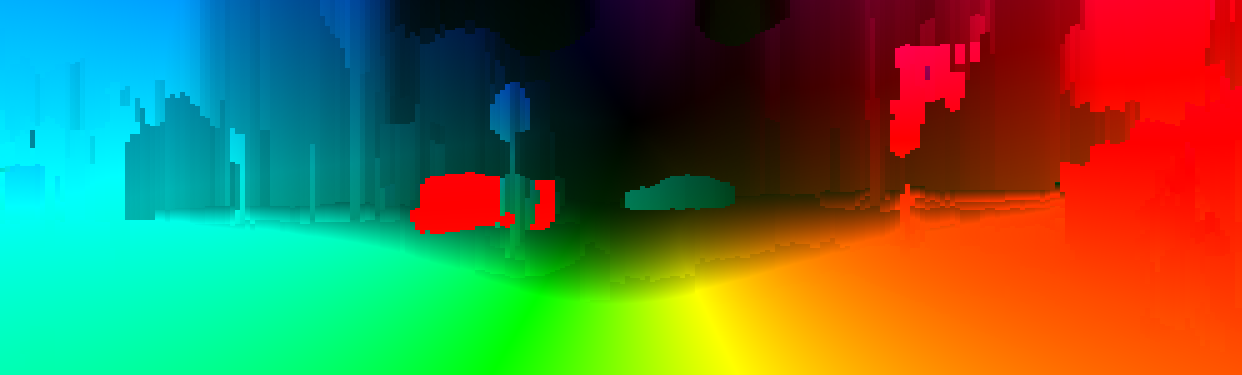} &
\includegraphics[width=0.188\textwidth]{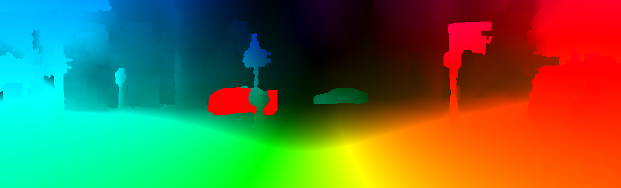} &
\raisebox{\normalbaselineskip}[0pt][0pt]{\rotatebox[origin=c]{-90}{\small Flow}}
\end{tabular}
\caption{Exemplary qualitative results of monocular scene flow estimation methods on the KITTI scene flow training set \cite{Menze2015CVPR}. 
  The top row of each scenario shows the depth values at time $t=0$ (left half) and $t=1$ (right half) colored from close (red) to far (dark blue). 
  The optical flow is visualized in the bottom row of each scenario.
  The ground truth is interpolated for visualization purposes.}
\label{fig_exampleResults}
\end{figure*}

%% file: images/ResultsCityscapes/Figure.tex
\begin{figure}[b]
\centering
\setlength\tabcolsep{1 pt}
\begin{tabular}{c c c}
\includegraphics[width=0.153\textwidth]{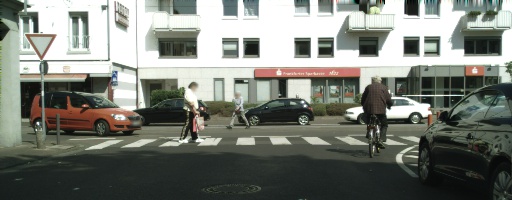} &
\includegraphics[width=0.153\textwidth]{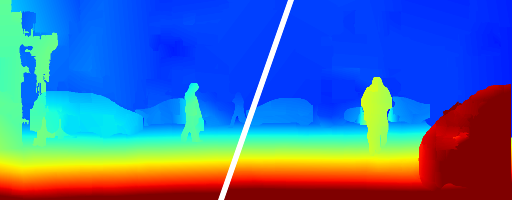} &
\includegraphics[width=0.153\textwidth]{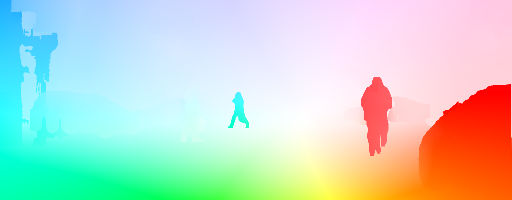} \\
\end{tabular}
\caption{Exemplary qualitative result of Mono-SF on a crop of Cityscapes (removing car hood);
 left: first input image, middle: estimated depth values at time $t=0$ (left half) and $t=1$ (right half),
 right: estimated optical flow}
\label{fig_exampleResultsCS}
\end{figure}

%% file: tables/Experiments/KittiSceneFlowTrain.tex
\begin{table*}[hbt]
\centering
\setlength\tabcolsep{2 pt}
\begin{tabular}{>{\centering}m{4.5cm} | c | c  c  c | c  c  c | c  c  c | c  c  c}
\multirow{2}{*}{Method} & \multirow{2}{*}{MRE} & \multicolumn{3}{c|}{D1} &  \multicolumn{3}{c|}{D2} &  \multicolumn{3}{c|}{Fl}   &  \multicolumn{3}{c}{SF}\\
& & bg & fg & all & bg & fg & all & bg & fg & all & bg & fg & all \\ \hline \hline
GeoNet \cite{yin2018geonet} & 20.08 & 47.03 & 63.41 & 49.54 & 56.25 & 68.82 & 58.17 & 32.43 & 67.69 & 37.83 & 67.69 & 91.41 & 71.32 \\
DF-Net \cite{zou2018df} & 18.95 & 44.43 & 57.94 & 46.50 & 61.55 & 61.47 & 61.54 & 25.66 & 37.45 & 27.47 & 71.63 & 82.52 & 73.30 \\
EveryPixel \cite{yang2018every} & - & 23.62 & 27.38 & 26.81 & - & - & - & 25.34 & 28.00 & 25.74 & - & - & - \\ \hline
MirrorFlow \cite{hur2017mirrorflow} + LRC \cite{godard2016deepdepth} & 9.06 & 25.33 & \textbf{19.83} & 24.49 & 35.83 & 26.15 & 34.34 & \textbf{9.40} & 14.22 & \textbf{10.14} & 40.55 & 35.17 & 39.73 \\ 
HD$^3$-F$^\dag$ \cite{yin2018hierarchical} + DORN$^\dag$ \cite{fu2018deep} & 11.18 & 17.02 & 37.54 & 20.16 & 30.08 & 40.47 & 31.67 & 4.01 & 6.76 & 4.43 & 32.57 & 46.89 & 34.76 \\
\hline
DMDE \cite{ranftl2016dense} & 14.6 & - & - & - & - & - & - & - & - & - & - & - & - \\ 
S. Soup \cite{kumar2017monocular} & 12.68 & - & - & - & - & - & - & - & - & - & - & - & - \\ \hline
MFA \cite{kumar2019motion} & 11.82 & - & - & - & - & - & - & - & - & - & - & - & - \\ 
Mono-Stixels \cite{monostixel2} & \textbf{8.04} & 18.28 & 22.06 & 18.86 & 22.00 & 31.19 & 23.41 & 9.84 & 14.36 & 10.54 & 24.03 & 39.13 & 26.34 \\
Mono-SF (ours) & 8.14 & \textbf{15.64} & 22.72 & \textbf{16.72} & \textbf{17.93} & \textbf{24.71} & \textbf{18.97} & 12.20 & \textbf{9.90} & 11.85 & \textbf{20.19} & \textbf{29.40} & \textbf{21.60} \\ \hline \hline
\multicolumn{14}{l}{\footnotesize \textit{ \textbf{MRE}: mean relative depth error at t=0 (capped at 50m); \textbf{D1} and \textbf{D2}: disparity errors at t=0,1; \textbf{Fl}: optical flow errors; \textbf{SF}: scene flow errors}} \\
\multicolumn{14}{l}{\footnotesize \textit{ \textbf{fg}: foreground (moving) ; \textbf{bg}: background (static); \textbf{all}: bg + fg ;\boldmath{$\dag$}: parts of dataset used for training (disregarded for ranking);} \textit{errors are in percent}}
\end{tabular}
\caption{ Quantitative evaluation of monocular scene flow methods on the KITTI scene flow training set \cite{MENZE201860}. 
  The methods are divided into four groups: 
  First, multi-task CNNs; 
  second, combining optical flow and single-view depth estimation as individual tasks; 
  third, multi-body or non-rigid SfM-based approaches;
  fourth, fusing single-view depth information with multi-view geometry.}
\label{tab_evaluationKittiTrain}
\end{table*}

%% file: tables/Experiments/KittiSceneFlowTest.tex
\begin{table}[t]
\centering
\setlength\tabcolsep{2 pt}
\begin{tabular}{>{\centering}m{4cm} | c c c c}
Method & D1-all & D2-all & Fl-all & SF-all \\ \hline \hline
UberATG-DSSF \cite{Ma2019CVPR} & 2.55 & 4.04 & 4.73 & 6.31 \\
ISF \cite{behl2017bounding} & 4.46 & 5.95 & 6.22 & 8.08\\
SGM \cite{hirschmuller2005accurate} + SF \cite{hornacek2014sphereflow} & 6.84 & 15.60 & 21.67 & 24.98 \\ \hline
Mono-SF & 16.32 & 19.59 & 12.77 & 23.08  \\ \hline \hline
\end{tabular}
\caption{Results of Mono-SF on the KITTI scene flow test set compared to some stereo-based scene flow estimation methods. 
}
\label{tab_resultsKittiTest}
\end{table}

%% file: tables/Experiments/AblationProbDepthNet.tex
\begin{table}[b]
\centering
\setlength\tabcolsep{2 pt}
\begin{tabular}{>{\centering}m{4cm} | c c  c  c}
Method & D1-all & D2-all & Fl-all & SF-all \\ \hline \hline
Mono-SF (LRC \cite{godard2016deepdepth}) & 22.36 & 26.29 & 15.10 & 30.96 \\ 
Mono-SF (w/o prob. depth) & 25.49 & 28.80 & 15.04 & 33.59 \\ 
Mono-SF (w/o recalib.)& 20.32 & 23.37 & 15.50 & 26.91 \\ 
Mono-SF & \textbf{16.72} & \textbf{18.97} & \textbf{11.85} & \textbf{21.60} \\ \hline \hline
\end{tabular}
\caption{Ablation study on ProbDepthNet for Mono-SF.
For integrating single-view depth information,
ProbDepthNet is more suitable than LRC for single-view depth estimation (improvement over "(LRC \cite{godard2016deepdepth})");
especially due to the importance of providing single-view depth estimates in a probabilistic
(improvement over "(w/o prob. depth)" ) and well-calibrated form (improvement over "(w/o recalib.)") for Mono-SF.}
\label{tab_ablationProbDepthNet}
\end{table}

%% file: tables/Experiments/AblationMonoSF.tex
\begin{table}[t]
\centering
\setlength\tabcolsep{5 pt}
\begin{tabular}{c c c | c c c c }
\multicolumn{3}{c|}{Energy terms} & \multicolumn{4}{c}{Results} \\
$\Phi^{pho} $ &  $\Phi^{svd}$ & $\Psi$ & D1-all &  D2-all &  Fl-all & SF-all \\ \hline \hline
- & - & - & 18.72 & 21.30 & 15.18 & 25.92 \\
\checkmark & - & - & 21.20 & 23.41 & 13.85 & 26.11\\
\checkmark & \checkmark & - & 18.65 & 21.10 & 13.31 & 23.67 \\
\checkmark & \checkmark & \checkmark & \textbf{16.72} & \textbf{18.97} & \textbf{11.85} & \textbf{21.60} \\ \hline \hline
\end{tabular}
\caption{Ablation study on Mono-SF approach.
Using the Mono-SF optimization improves the scene flow estimation compared to its initialization (denoted by the row without checkmark).
Each term of the energy minimization problem (photometric distance($\Phi^{pho}$), single-view depth ($\Phi^{svd}$) and smoothness prior ($\Psi$)) 
contributes to the final performance.}
\label{tab_ablationMonoSF}
\end{table}

%% file: text/Conclusion/Conclusion.tex
In this paper, we proposed Mono-SF for joint estimation of the 3D geometry and motion of particularly traffic scenes 
by combining multi-view geometry with single-view depth information.
For a sensible statistical integration, we showed the importance of providing 
 single-view depth information in a probabilistic and well-calibrated form,
 which is made possible by our proposed ProbDepthNet including a novel recalibration technique.

%% file: ms.bbl
\begin{thebibliography}{10}\itemsep=-1pt

\bibitem{bai2016exploiting}
Min Bai, Wenjie Luo, Kaustav Kundu, and Raquel Urtasun.
\newblock {Exploiting semantic information and deep matching for optical flow}.
\newblock In {\em Proc. of European Conference on Computer Vision (ECCV)},
  pages 154--170. Springer, 2016.

\bibitem{barnes2018driven}
Dan Barnes, Will Maddern, Geoffrey Pascoe, and Ingmar Posner.
\newblock {Driven to distraction: Self-supervised distractor learning for
  robust monocular visual odometry in urban environments}.
\newblock In {\em Proc. of IEEE International Conference on Robotics and
  Automation (ICRA)}, pages 1894--1900. IEEE, 2018.

\bibitem{basha2013multi}
Tali Basha, Yael Moses, and Nahum Kiryati.
\newblock {Multi-view scene flow estimation: A view centered variational
  approach}.
\newblock {\em {International Journal of Computer Vision}}, 101(1):6--21, 2013.

\bibitem{behl2017bounding}
Aseem Behl, Omid~Hosseini Jafari, Siva~Karthik Mustikovela, Hassan~Abu Alhaija,
  Carsten Rother, and Andreas Geiger.
\newblock {Bounding Boxes, Segmentations and Object Coordinates: How Important
  is Recognition for 3D Scene Flow Estimation in Autonomous Driving Scenarios?}
\newblock In {\em Proc. of IEEE Conference on Computer Vision and Pattern
  Recognition (CVPR)}, pages 2574--2583, 2017.

\bibitem{monostixel2}
Fabian Brickwedde, Steffen Abraham, and Rudolf Mester.
\newblock {Exploiting Single Image Depth Prediction for Mono-Stixel
  Estimation}.
\newblock In {\em Proc. of European Conference of Computer Vision Workshops
  (ECCV Workshops)}. IEEE, 2018.

\bibitem{monostixel}
Fabian Brickwedde, Steffen Abraham, and Rudolf Mester.
\newblock {{Mono-Stixels}: {Monocular Depth Reconstruction of Dynamic Street
  Scenes}}.
\newblock In {\em Proc. of IEEE International Conference on Robotics and
  Automation (ICRA)}, pages 1--7. IEEE, 2018.

\bibitem{brikbeck2010depth}
Neil Brikbeck, Dana Cobzas, and Martin J{\"a}gersand.
\newblock {Depth and scene flow from a single moving camera}.
\newblock In {\em Proc. of International Symposium on 3D Data Processing,
  Visualization and Transmission (3DPVT)}, 2010.

\bibitem{bullinger20183d}
Sebastian Bullinger, Christoph Bodensteiner, Michael Arens, and Rainer
  Stiefelhagen.
\newblock {3D Vehicle Trajectory Reconstruction in Monocular Video Data Using
  Environment Structure Constraints}.
\newblock In {\em Proc. of European Conference on Computer Vision (ECCV)},
  pages 35--50, 2018.

\bibitem{cordts2016cityscapes}
Marius Cordts, Mohamed Omran, Sebastian Ramos, Timo Rehfeld, Markus Enzweiler,
  Rodrigo Benenson, Uwe Franke, Stefan Roth, and Bernt Schiele.
\newblock {The cityscapes dataset for semantic urban scene understanding}.
\newblock In {\em Proc. of the IEEE Conference on Computer Vision and Pattern
  Recognition (CVPR)}, pages 3213--3223, 2016.

\bibitem{eigen2014depth}
David Eigen, Christian Puhrsch, and Rob Fergus.
\newblock {Depth map prediction from a single image using a multi-scale deep
  network}.
\newblock In {\em Proc. of Advances in neural information processing systems
  (NeurIPS)}, pages 2366--2374, 2014.

\bibitem{engel2014lsd}
Jakob Engel, Thomas Sch{\"o}ps, and Daniel Cremers.
\newblock {{LSD-SLAM}: Large-scale direct monocular {SLAM}}.
\newblock In {\em Proc. of European Conference on Computer Vision (ECCV)},
  pages 834--849. Springer, 2014.

\bibitem{facil2017fusion}
{F{\'a}cil, Jos{\'e} M and Concha, Alejo and Montesano, Luis and Civera,
  Javier}.
\newblock {Single-View and Multi-View Depth Fusion}.
\newblock {\em IEEE Robotics and Automation Letters}, 2(4):1994--2001, Oct
  2017.

\bibitem{fananipmo}
Nolang Fanani, Alina St{\"u}rck, Matthias Ochs, Henry Bradler, and Rudolf
  Mester.
\newblock {Predictive monocular odometry ({PMO}): What is possible without
  {RANSAC} and multiframe bundle adjustment?}
\newblock {\em Image and Vision Computing}, 2017.

\bibitem{fu2018deep}
Huan Fu, Mingming Gong, Chaohui Wang, Kayhan Batmanghelich, and Dacheng Tao.
\newblock {Deep Ordinal Regression Network for Monocular Depth Estimation}.
\newblock In {\em Proc. of the IEEE Conference on Computer Vision and Pattern
  Recognition (CVPR)}, pages 2002--2011, 2018.

\bibitem{garg2016singleviewdepth}
Ravi Garg, Gustavo Carneiro, and Ian Reid.
\newblock {Unsupervised {CNN} for single view depth estimation: Geometry to the
  rescue}.
\newblock In {\em Proc. of European Conference on Computer Vision (ECCV)},
  pages 740--756. Springer, 2016.

\bibitem{garg2013dense}
Ravi Garg, Anastasios Roussos, and Lourdes Agapito.
\newblock {Dense variational reconstruction of non-rigid surfaces from
  monocular video}.
\newblock In {\em Proc. of IEEE Conference on Computer Vision and Pattern
  Recognition (CVPR)}, pages 1272--1279, 2013.

\bibitem{gast2018lightweight}
Jochen Gast and Stefan Roth.
\newblock {Lightweight Probabilistic Deep Networks}.
\newblock In {\em Proc. of IEEE Conference on Computer Vision and Pattern
  Recognition (CVPR)}, pages 3369--3378, 2018.

\bibitem{geiger2011stereoscan}
Andreas Geiger, Julius Ziegler, and Christoph Stiller.
\newblock {Stereoscan: Dense {3D} reconstruction in real-time}.
\newblock In {\em Proc. of IEEE Intelligent Vehicles Symposium (IV)}, pages
  963--968, 2011.

\bibitem{godard2016deepdepth}
Clement Godard, Oisin Mac~Aodha, and Gabriel~J. Brostow.
\newblock {Unsupervised Monocular Depth Estimation With Left-Right
  Consistency}.
\newblock In {\em Proc. of IEEE Conference on Computer Vision and Pattern
  Recognition (CVPR)}, July 2017.

\bibitem{golyanik2016nrsfm}
Vladislav Golyanik, Aman~S Mathur, and Didier Stricker.
\newblock {NRSfM-Flow: Recovering Non-Rigid Scene Flow from Monocular Image
  Sequences.}
\newblock In {\em Proc. of British Machine Vision Conference (BMVC)}, 2016.

\bibitem{guo2017calibration}
Chuan Guo, Geoff Pleiss, Yu Sun, and Kilian~Q Weinberger.
\newblock {On Calibration of Modern Neural Networks}.
\newblock In {\em Proc. of International Conference on Machine Learning
  (ICML)}, pages 1321--1330, 2017.

\bibitem{hartley2003multiple}
Richard Hartley and Andrew Zisserman.
\newblock {\em {Multiple view geometry in computer vision}}.
\newblock Cambridge university press, 2003.

\bibitem{he2017mask}
Kaiming He, Georgia Gkioxari, Piotr Doll{\'a}r, and Ross Girshick.
\newblock {Mask R-CNN}.
\newblock In {\em Proc. of IEEE International Conference on Computer Vision
  (ICCV)}, pages 2980--2988. IEEE, 2017.

\bibitem{he2016deep}
Kaiming He, Xiangyu Zhang, Shaoqing Ren, and Jian Sun.
\newblock {Deep residual learning for image recognition}.
\newblock In {\em Proc. of the IEEE Conference on Computer Vision and Pattern
  Recognition (CVPR)}, pages 770--778, 2016.

\bibitem{herbst2013rgbdflow}
Evan Herbst, Xiaofeng Ren, and Dieter Fox.
\newblock {RGB-D flow: Dense 3-D motion estimation using color and depth}.
\newblock In {\em Proc. of IEEE International Conference on Robotics and
  Automation (ICRA)}, pages 2276--2282, May 2013.

\bibitem{hirschmuller2005accurate}
Heiko Hirschmuller.
\newblock {Accurate and efficient stereo processing by semi-global matching and
  mutual information}.
\newblock In {\em Proc. of IEEE Conference on Computer Vision and Pattern
  Recognition (CVPR)}, pages 807--814, 2005.

\bibitem{hoiem2005automatic}
Derek Hoiem, Alexei~A Efros, and Martial Hebert.
\newblock {Automatic photo pop-up}.
\newblock In {\em Proc. of ACM transactions on graphics (TOG)}, volume~24,
  pages 577--584. ACM, 2005.

\bibitem{hornacek2014sphereflow}
Michael Hornacek, Andrew Fitzgibbon, and Carsten Rother.
\newblock {SphereFlow: 6 DoF scene flow from RGB-D pairs}.
\newblock In {\em Proc. of IEEE Conference on Computer Vision and Pattern
  Recognition (CVPR)}, pages 3526--3533, 2014.

\bibitem{huguet2007variational}
Fr{\'e}d{\'e}ric Huguet and Fr{\'e}d{\'e}ric Devernay.
\newblock {A variational method for scene flow estimation from stereo
  sequences}.
\newblock In {\em Proc. of IEEE International Conference on Computer Vision},
  pages 1--7. IEEE, 2007.

\bibitem{hur2017mirrorflow}
Junhwa Hur and Stefan Roth.
\newblock {MirrorFlow: Exploiting symmetries in joint optical flow and
  occlusion estimation}.
\newblock In {\em Proc. of International Conference on Computer Vision (ICCV)},
  2017.

\bibitem{ICKMB18}
Eddy Ilg, Ozgun Cicek, Silvio Galesso, Aaron Klein, Osama Makansi, Frank
  Hutter, and Thomas Brox.
\newblock {Uncertainty Estimates and Multi-Hypotheses Networks for Optical
  Flow}.
\newblock In {\em Proc. of European Conference on Computer Vision (ECCV)},
  2018.

\bibitem{kendall2017uncertainties}
Alex Kendall and Yarin Gal.
\newblock {What uncertainties do we need in bayesian deep learning for computer
  vision?}
\newblock In {\em Proc. of Advances in neural information processing systems
  (NeurIPS)}, pages 5574--5584, 2017.

\bibitem{kingma2014adam}
Diederik~P Kingma and Jimmy~Lei Ba.
\newblock {Adam: Amethod for stochastic optimization}.
\newblock In {\em Proc. of International Conference for Learning
  Representations (ICLR)}, 2014.

\bibitem{klodt2018supervising}
Maria Klodt and Andrea Vedaldi.
\newblock {Supervising the new with the old: learning SFM from SFM}.
\newblock In {\em Proc. of European Conference on Computer Vision (ECCV)},
  pages 698--713, 2018.

\bibitem{kuleshov2018accurate}
Volodymyr Kuleshov, Nathan Fenner, and Stefano Ermon.
\newblock {Accurate Uncertainties for Deep Learning Using Calibrated
  Regression}.
\newblock In {\em Proc. of International Conference on Machine Learning
  (ICML)}, pages 2801--2809, 2018.

\bibitem{kumar2017monocular}
Suryansh Kumar, Yuchao Dai, and Hongdong Li.
\newblock {Monocular dense 3D reconstruction of a complex dynamic scene from
  two perspective frames}.
\newblock In {\em Proc. of IEEE International Conference on Computer Vision
  (ICCV)}, pages 4649--4657, 2017.

\bibitem{kumar2019motion}
Suryansh Kumar, Ram~Srivatsav Ghorakavi, Yuchao Dai, and Hongdong Li.
\newblock {A Motion Free Approach to Dense Depth Estimation in Complex Dynamic
  Scene}.
\newblock {\em arXiv preprint arXiv:1902.03791}, 2019.

\bibitem{Kuemmerle2011}
Rainer K{\"u}mmerle, Giorgio Grisetti, Hauke Strasdat, Kurt Konolige, and
  Wolfram Burgard.
\newblock {g 2 o: A general framework for graph optimization}.
\newblock In {\em Proc. of IEEE International Conference on Robotics and
  Automation (ICRA)}, pages 3607--3613. IEEE, 2011.

\bibitem{kuznietsov2017semi}
Yevhen Kuznietsov, J{\"o}rg St{\"u}ckler, and Bastian Leibe.
\newblock {Semi-supervised deep learning for monocular depth map prediction}.
\newblock In {\em Proc. of the IEEE Conference on Computer Vision and Pattern
  Recognition (CVPR)}, pages 6647--6655, 2017.

\bibitem{ladicky2014pulling}
Lubor Ladicky, Jianbo Shi, and Marc Pollefeys.
\newblock {Pulling things out of perspective}.
\newblock In {\em Proceedings of the IEEE Conference on Computer Vision and
  Pattern Recognition (CVPR)}, pages 89--96, 2014.

\bibitem{Ma2019CVPR}
Wei-Chiu Ma, Shenlong Wang, Rui Hu, Yuwen Xiong, and Raquel Urtasun.
\newblock Deep rigid instance scene flow.
\newblock In {\em Proc. of IEEE Conference on Computer Vision and Pattern
  Recognition (CVPR)}, June 2019.

\bibitem{mahjourian2018unsupervised}
Reza Mahjourian, Martin Wicke, and Anelia Angelova.
\newblock {Unsupervised Learning of Depth and Ego-Motion from Monocular Video
  Using 3D Geometric Constraints}.
\newblock In {\em Proc. of IEEE Conference on Computer Vision and Pattern
  Recognition (CVPR)}, pages 5667--5675, 2018.

\bibitem{malinin2018predictive}
Andrey Malinin and Mark Gales.
\newblock Predictive uncertainty estimation via prior networks.
\newblock In {\em Proc. of Advances in Neural Information Processing Systems
  (NeurIPS)}, pages 7047--7058, 2018.

\bibitem{Menze2015CVPR}
Moritz Menze and Andreas Geiger.
\newblock {Object Scene Flow for Autonomous Vehicles}.
\newblock In {\em Proc. of IEEE Conference on Computer Vision and Pattern
  Recognition (CVPR)}, 2015.

\bibitem{MENZE201860}
Moritz Menze, Christian Heipke, and Andreas Geiger.
\newblock {Object Scene Flow}.
\newblock {\em ISPRS Journal of Photogrammetry and Remote Sensing}, 140:60 --
  76, 2018.
\newblock Geospatial Computer Vision.

\bibitem{mitiche15mono}
Amar Mitiche, Yosra Mathlouthi, and Ismail Ben~Ayed.
\newblock {Monocular Concurrent Recovery of Structure and Motion Scene Flow}.
\newblock {\em Frontiers in ICT}, 2:16, 2015.

\bibitem{mur2015orb}
Raul Mur-Artal, Jose Maria~Martinez Montiel, and Juan~D Tardos.
\newblock {{ORB-SLAM}: a versatile and accurate monocular {SLAM} system}.
\newblock {\em IEEE Transactions on Robotics}, 31(5):1147--1163, 2015.

\bibitem{pereira2017monocular}
Fabio~Irigon Pereira, Gustavo Ilha, Joel Luft, Marcelo Negreiros, and Altamiro
  Susin.
\newblock {Monocular Visual Odometry with Cyclic Estimation}.
\newblock In {\em Graphics, Patterns and Images (SIBGRAPI), 2017 30th SIBGRAPI
  Conference on}, pages 1--6. IEEE, 2017.

\bibitem{platt1999probabilistic}
John Platt.
\newblock {Probabilistic outputs for support vector machines and comparisons to
  regularized likelihood methods}.
\newblock {\em Advances in large margin classifiers}, 10(3):61--74, 1999.

\bibitem{pons2007multi}
Jean-Philippe Pons, Renaud Keriven, and Olivier Faugeras.
\newblock {Multi-view stereo reconstruction and scene flow estimation with a
  global image-based matching score}.
\newblock {\em International Journal of Computer Vision}, 72(2):179--193, 2007.

\bibitem{ranftl2016dense}
Ren{\'e} Ranftl, Vibhav Vineet, Qifeng Chen, and Vladlen Koltun.
\newblock {Dense monocular depth estimation in complex dynamic scenes}.
\newblock In {\em Proc. of IEEE Conference on Computer Vision and Pattern
  Recognition (CVPR)}, pages 4058--4066, 2016.

\bibitem{saxena2005learning}
Ashutosh Saxena, Sung~H Chung, and Andrew~Y Ng.
\newblock {Learning depth from single monocular images}.
\newblock In {\em Proc. of Advances in Neural Information Processing Systems
  (NeurIPS)}, volume~18, pages 1--8, 2005.

\bibitem{saxenamake3d}
Ashutosh Saxena, Min Sun, and Andrew~Y Ng.
\newblock {Make3D: Learning {3D} Scene Structure from a Single Still Image}.
\newblock {\em IEEE Transactions on Pattern Analysis and Machine Intelligence},
  31(5):824--840, May 2009.

\bibitem{tateno2017cnn}
Keisuke Tateno, Federico Tombari, Iro Laina, and Nassir Navab.
\newblock {{CNN-SLAM}: Real-time dense monocular {SLAM} with learned depth
  prediction}.
\newblock In {\em Proc. of the IEEE Conference on Computer Vision and Pattern
  Recognition (CVPR)}, volume~2, 2017.

\bibitem{teng2018occlusion}
Qianru Teng, Yimin Chen, and Chen Huang.
\newblock {Occlusion-Aware Unsupervised Learning of Monocular Depth, Optical
  Flow and Camera Pose with Geometric Constraints}.
\newblock {\em Future Internet}, 10(10):92, 2018.

\bibitem{uhrig2017sparsity}
Jonas Uhrig, N Schneider, L Schneider, U Franke, Thomas Brox, and A Geiger.
\newblock {Sparsity Invariant CNNs}.
\newblock In {\em Proc. of IEEE International Conference on 3D Vision (3DV)},
  2017.

\bibitem{ummenhofer2017demon}
Benjamin Ummenhofer, Huizhong Zhou, Jonas Uhrig, Nikolaus Mayer, Eddy Ilg,
  Alexey Dosovitskiy, and Thomas Brox.
\newblock {{DeMoN}: Depth and Motion Network for Learning Monocular Stereo}.
\newblock In {\em Proc. of IEEE Conference on Computer Vision and Pattern
  Recognition (CVPR)}, July 2017.

\bibitem{valgaerts2010joint}
Levi Valgaerts, Andr{\'e}s Bruhn, Henning Zimmer, Joachim Weickert, Carsten
  Stoll, and Christian Theobalt.
\newblock {Joint estimation of motion, structure and geometry from stereo
  sequences}.
\newblock In {\em Proc. of European Conference on Computer Vision (ECCV)},
  pages 568--581. Springer, 2010.

\bibitem{vedula1999three}
Sundar Vedula, Simon Baker, Peter Rander, Robert Collins, and Takeo Kanade.
\newblock {Three-dimensional scene flow}.
\newblock In {\em Proc. of IEEE International Conference on Computer Vision
  (ICCV)}, volume~2, pages 722--729. IEEE, 1999.

\bibitem{vedula2005three}
Sundar Vedula, Peter Rander, Robert Collins, and Takeo Kanade.
\newblock {Three-dimensional scene flow}.
\newblock {\em {IEEE transactions on pattern analysis and machine
  intelligence}}, 27(3):475--480, 2005.

\bibitem{vogel2013piecewise}
Christoph Vogel, Konrad Schindler, and Stefan Roth.
\newblock {Piecewise rigid scene flow}.
\newblock In {\em Proc. of the IEEE International Conference on Computer Vision
  (ICCV)}, pages 1377--1384, 2013.

\bibitem{wang2018learning}
Chaoyang Wang, Jos{\'e}~Miguel Buenaposada, Rui Zhu, and Simon Lucey.
\newblock {Learning Depth from Monocular Videos using Direct Methods}.
\newblock In {\em Proc. of IEEE Conference on Computer Vision and Pattern
  Recognition (CVPR)}, pages 2022--2030, 2018.

\bibitem{mxmaskrcnn}
Naiyan Wang.
\newblock {An MXNet implementation of Mask R-CNN}.
\newblock https://github.com/TuSimple/mx-maskrcnn, 2018.
\newblock [accessed June, 25 2018].

\bibitem{wedel2011stereoscopic}
Andreas Wedel, Thomas Brox, Tobi Vaudrey, Clemens Rabe, Uwe Franke, and Daniel
  Cremers.
\newblock {Stereoscopic scene flow computation for 3D motion understanding}.
\newblock {\em International Journal of Computer Vision}, 95(1):29--51, 2011.

\bibitem{wedel2008efficient}
Andreas Wedel, Clemens Rabe, Tobi Vaudrey, Thomas Brox, Uwe Franke, and Daniel
  Cremers.
\newblock {Efficient dense scene flow from sparse or dense stereo data}.
\newblock In {\em Proc. of European Conference on Computer Vision (ECCV)},
  pages 739--751. Springer, 2008.

\bibitem{Xiao2017}
Degui Xiao, Qiuwei Yang, Bing Yang, and Wei Wei.
\newblock {Monocular scene flow estimation via variational method}.
\newblock {\em Multimedia Tools and Applications}, 76(8):10575--10597, Apr
  2017.

\bibitem{yamaguchi2013robust}
Koichiro Yamaguchi, David McAllester, and Raquel Urtasun.
\newblock {Robust monocular epipolar flow estimation}.
\newblock In {\em Proc. of IEEE Conference on Computer Vision and Pattern
  Recognition (CVPR)}, pages 1862--1869, 2013.

\bibitem{sps}
Koichiro Yamaguchi, David McAllester, and Raquel Urtasun.
\newblock {Efficient Joint Segmentation, Occlusion Labeling, Stereo and Flow
  Estimation}.
\newblock In David Fleet, Tomas Pajdla, Bernt Schiele, and Tinne Tuytelaars,
  editors, {\em Proc. of European Conference on Computer Vision (ECCV)}, pages
  756--771, Cham, 2014. Springer International Publishing.

\bibitem{yang2018deep}
Nan Yang, Rui Wang, J{\"o}rg St{\"u}ckler, and Daniel Cremers.
\newblock {Deep virtual stereo odometry: Leveraging deep depth prediction for
  monocular direct sparse odometry}.
\newblock In {\em Proc. of European Conference on Computer Vision (ECCV)},
  pages 835--852. Springer, 2018.

\bibitem{yang2018every}
Zhenheng Yang, Peng Wang, Yang Wang, Wei Xu, and Ram Nevatia.
\newblock {Every Pixel Counts: Unsupervised Geometry Learning with Holistic 3D
  Motion Understanding}.
\newblock In {\em Proc. of European Conference on Computer Vision Workshops
  (ECCV Workshops)}, 2018.

\bibitem{yin2017scale}
Xiaochuan Yin, Xiangwei Wang, Xiaoguo Du, and Qijun Chen.
\newblock Scale recovery for monocular visual odometry using depth estimated
  with deep convolutional neural fields.
\newblock In {\em Proceedings of the IEEE International Conference on Computer
  Vision}, pages 5870--5878, 2017.

\bibitem{yin2018hierarchical}
Zhichao Yin, Trevor Darrell, and Fisher Yu.
\newblock Hierarchical discrete distribution decomposition for match density
  estimation.
\newblock In {\em Proc. of IEEE Conference on Computer Vision and Pattern
  Recognition (CVPR)}, June 2019.

\bibitem{yin2018geonet}
Zhichao Yin and Jianping Shi.
\newblock {{GeoNet: Unsupervised Learning of Dense Depth, Optical Flow and
  Camera Pose}}.
\newblock In {\em Proc. of IEEE Conference on Computer Vision and Pattern
  Recognition (CVPR)}, volume~2, 2018.

\bibitem{zabih1994non}
Ramin Zabih and John Woodfill.
\newblock {Non-parametric local transforms for computing visual
  correspondence}.
\newblock In {\em Proc. of European Conference on Computer Vision (ECCV)},
  pages 151--158. Springer, 1994.

\bibitem{zhan2018unsupervised}
Huangying Zhan, Ravi Garg, Chamara~Saroj Weerasekera, Kejie Li, Harsh Agarwal,
  and Ian Reid.
\newblock {Unsupervised Learning of Monocular Depth Estimation and Visual
  Odometry with Deep Feature Reconstruction}.
\newblock In {\em Proc. of IEEE Conference on Computer Vision and Pattern
  Recognition (CVPR)}, pages 340--349, 2018.

\bibitem{zhou2017unsupervised}
Tinghui Zhou, Matthew Brown, Noah Snavely, and David~G. Lowe.
\newblock {Unsupervised Learning of Depth and Ego-Motion from Video}.
\newblock In {\em Proc. of IEEE Conference on Computer Vision and Pattern
  Recognition (CVPR)}, 2017.

\bibitem{zou2018df}
Yuliang Zou, Zelun Luo, and Jia-Bin Huang.
\newblock {DF-Net: Unsupervised Joint Learning of Depth and Flow using
  Cross-Network Consistency}.
\newblock In {\em Proc. of European Conference on Computer Vision (ECCV)},
  pages 36--53, 2018.

\end{thebibliography}
